\documentclass[10pt, journal, a4paper, final, twoside]{IEEEtran}
\usepackage{nicematrix, cite,graphicx,subfigure,mhchem,amsmath, mathtools}
\usepackage{amssymb}
\usepackage{color, colortbl}
\usepackage[english]{babel}
\usepackage[top=1.5cm,bottom=1.5cm,left=1.5cm,right=1.5cm,marginparwidth=1.75cm]{geometry}
\usepackage[colorlinks=true, allcolors=blue]{hyperref}
\usepackage{times}
\usepackage{dblfloatfix}
\usepackage{booktabs}
\usepackage{longtable}
\usepackage{array}
\usepackage{enumitem}
\usepackage{multirow}

\usepackage[most]{tcolorbox} 
\usepackage{tcolorbox}
\tcbuselibrary{skins, breakable} 

\usepackage{tabularx}
\usepackage[table,x11names]{xcolor} 
\usepackage{caption}

\newcommand{\neutral}[1]{\pagebreak[0]\cellcolor{gray!10!white} #1}

\definecolor{biotechblue}{HTML}{1A5276}
\definecolor{lightbg}{HTML}{F4F6F7}

\title{Closed-Loop LLM Co-Pilots for Digital Agriculture}
\author{ \textbf{Serge Kernbach}$^1$
\vspace{-8mm}
\thanks{$^1$ CYBRES GmbH, Research Center of Advanced Robotics and Environmental Science, Melunerstr. 40, 70569 Stuttgart, Germany, serge.kernbach@cybertronica.de.com}
}
\date{}

\begin{document}

\maketitle

\begin{abstract}
This study evaluates the application of Large Language Models (LLMs) in complex biological systems, evolving from data analysis to autonomous, AI-guided experimentation. The framework is driven by data from a 49-channel phytosensor network, using hydrodynamic transport and photochemical reflectance models to evaluate multispectral, electrochemical, and dielectric parameters. To enhance accessibility, the system provides real-time natural-language interpretation for both specialists and non-experts. However, its core advantage lies in the transition from human-in-the-loop analysis to fully autonomous control. Processing biophysical data, the LLM evaluates plant physiology and triggers hardware actuators to optimize microclimates, execute phenotyping protocols, or induce controlled stress scenarios. This closed-loop architecture establishes a direct AI-biology interface, enabling data-driven exploration of complex biosystems and ecologies. The framework was validated across three case studies, based on a vertical farm and a single-plant setup and deciphered complex micro- and macro-fluctuations in plant physiology that elude manual interpretation. Agents in a production-scale deployment executed multi-parameter optimization, balancing biomass accumulation, chlorophyll content, and energy consumption. The LLM processed biosensing telemetry to modulate full-spectrum, 450~nm, and 660~nm lighting at 2-hour intervals over a 24-hour cycle. Compared to periodic control, the system in minimal-time mode reduced the production cycle by 35\%. In the energy-optimization mode, it reduced energy consumption by 18\% with only a marginal increase in cultivation time, exploiting physiological inertia via short light pulses. Finally, the agents autonomously developed an unforeseen strategy of dark-induced chlorophyll accumulation, resulting in a 67.9\% energy saving. The integration of AI into engineering and biological R\&D significantly advances the development of bio-hybrid systems; however, it introduces critical challenges in interpretability. This framework transforms LLMs into autonomous co-pilots for digital agriculture, improving the cost-to-value ratio and lowering computational and expert-labor constraints.
 
\begin{figure}[h]
\centering
\subfigure{\includegraphics[width=.47\textwidth]{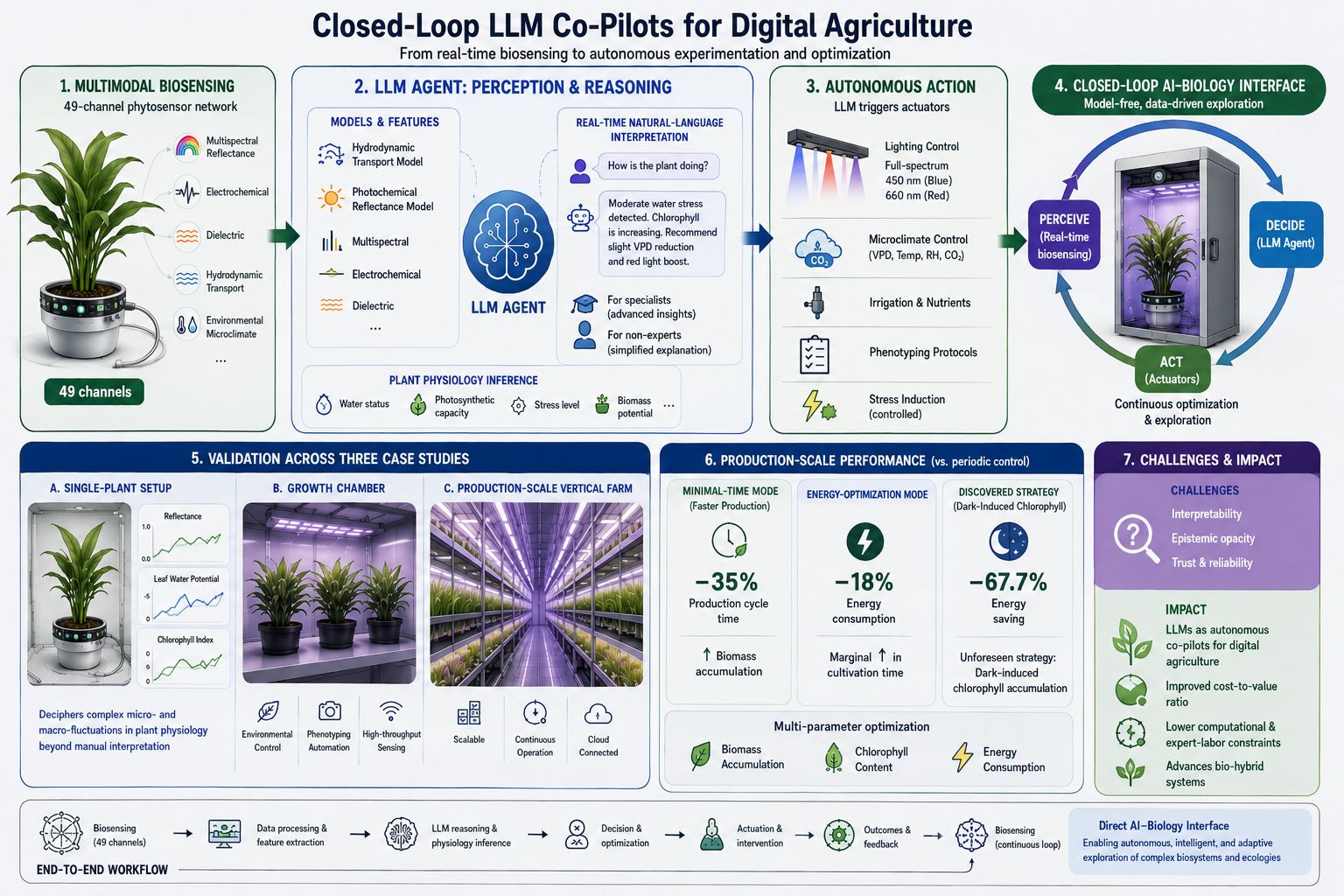}}
\end{figure}

\end{abstract}


\section{Introduction}

Modern digital agriculture platforms and automated plant phenotype screening generate large volumes of multi-channel time-series data. With continuous monitoring running at high sampling frequencies across multiple sensor grids, researchers, agronomists, and farmers confront the "data rich, information poor" dilemma \cite{wolfert2017big}. Beyond data interpretation, multi-parameter datasets introduce optimal control challenges due to the nonlinear, time-variant, and stochastic nature of plant physiological responses \cite{van2011optimal}.

To address these issues, this study evaluates a human-AI framework where an LLM acts as an interactive co-pilot within the data-processing loop. The core utility of LLMs originates from their capacity for cross-domain cognitive synthesis \cite{10.1007/s13748-024-00359-4}, thereby overcoming the limitations of manual analysis. These models recognize non-linear sensor alignments across different modalities -- such as correlations between stem bio-impedance, stomatal regulation, and ambient air pollutants \cite{kernbach2026Ozone}. This cognitive capacity enables real-time translation of complex biophysical data into physiological hypotheses, delivered in natural language to assist both specialists and non-experts.

In addition to real-time analytics, the LLM functions as a controller within a closed-loop cyber-physical system. In this architecture, the model parses biophysical data in real time \cite{wang2024survey} and translates these inputs into automated commands. Through integration with hardware phyto-actuators, the LLM directly manipulates the environment based on these real-time plant responses \cite{Vemprala2023ChatGPTFR}. This control loop is non-trivial; rather than executing static rules, the agent introduces cognitive intelligence to evaluate dependencies between plant metabolism, ecology, and resource consumption. This closed-loop configuration establishes a direct AI-biology interface, enabling an adaptive, data-driven approach to exploring complex biosystems and ecologies, which represents the primary novelty of this work.

The framework was validated using a real-time data stream from two setups: a vertical farm producing microgreens (wheatgrass, peas) and individual plants (Dracaena, tomato, pepper). Utilizing a 49-channel phytomonitoring network with electrochemical, dielectric, and optical spectroscopy \cite{kernbach2024Biohybrid}, the system condensed raw data into 12 to 15 integrated parameters derived from biological models \cite{10.1093/jxb/erq018}. The AI evaluated analytical schemes to optimize data aggregation for specific agricultural or physiological tasks \cite{Buss23}. The first case study investigated a closed-loop feedback architecture where the LLMs optimized full, 450~nm, and 660~nm spectral intensities to maximize biomass accumulation or minimize energy consumption. The system re-evaluated actuation steps every two hours to maintain 24-hour autonomous operation. In this configuration, the agents functioned as a data-driven meta-controller responding directly to physiological changes. 

Two other case studies evaluated the capacity of agents to bridge the gap between raw biological data and its physiological meaning. Demonstrated through the analysis of water balance disruptions and microclimatic stress, these methods applied to long-term trend evaluation in vertical farms or standalone monitoring for individual plants. The distinction between these cases lies in the feature engineering stage: the agents either generated and ran Python scripts to aggregate data or processed small segments directly.

The primary objective of this work is to demonstrate the utility of LLMs for rapid plant diagnostics, bypassing labor-intensive manual modeling. Furthermore, this architecture enables real-time optimal control of dynamic biosystems through adaptive strategies. We evaluate the interoperability gap and epistemic opacity caused by the high complexity of biological data. This approach lowers the deployment and computational barriers of phytosensing infrastructure, improving the cost-to-value ratio of automated agricultural monitoring.

\section{Materials and Methods}

\begin{figure}[h]
\centering
\subfigure[\label{fig:setupA}]{\includegraphics[width=.48\textwidth]{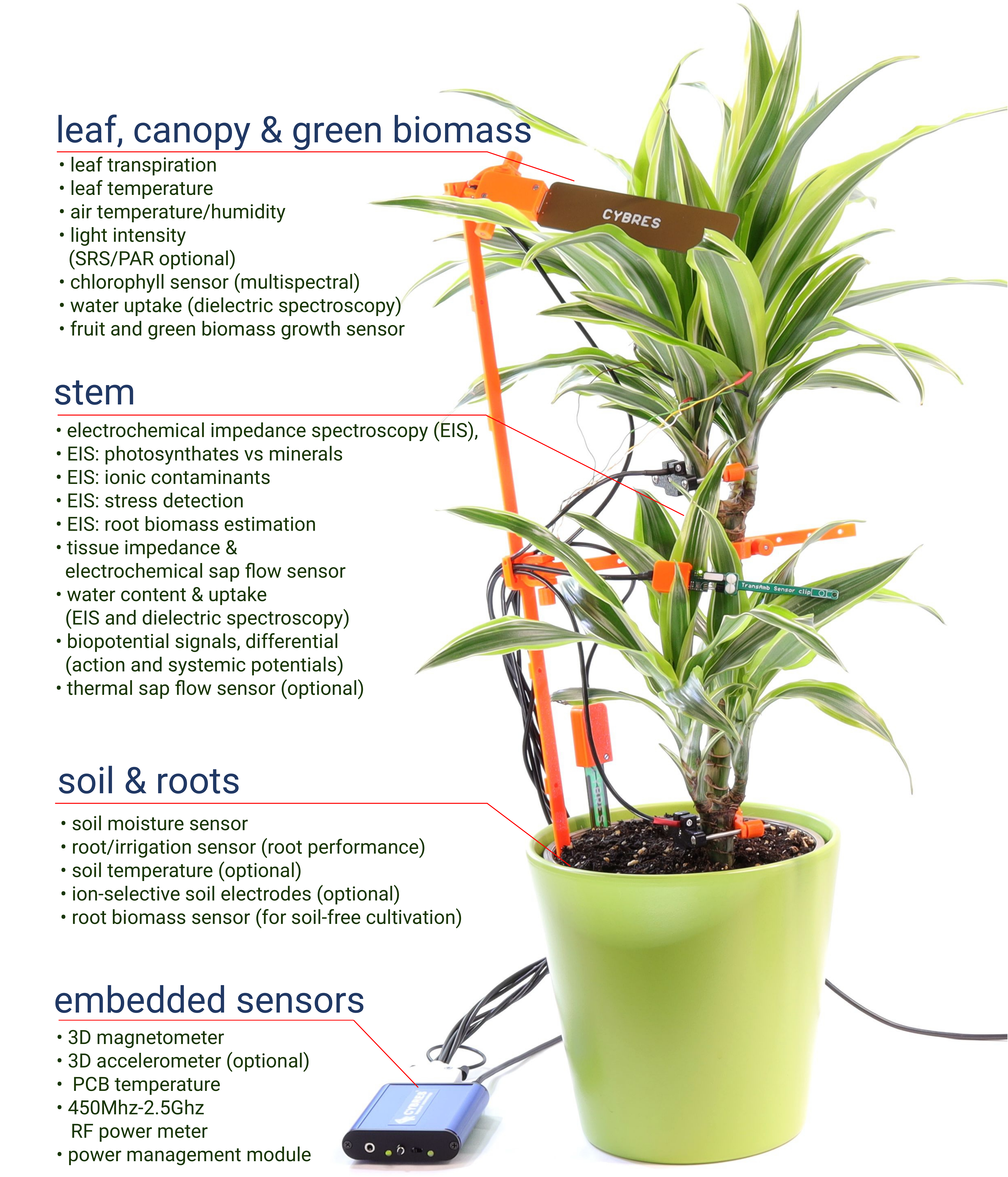}}
\subfigure[\label{fig:setupB}]{\includegraphics[width=.48\textwidth]{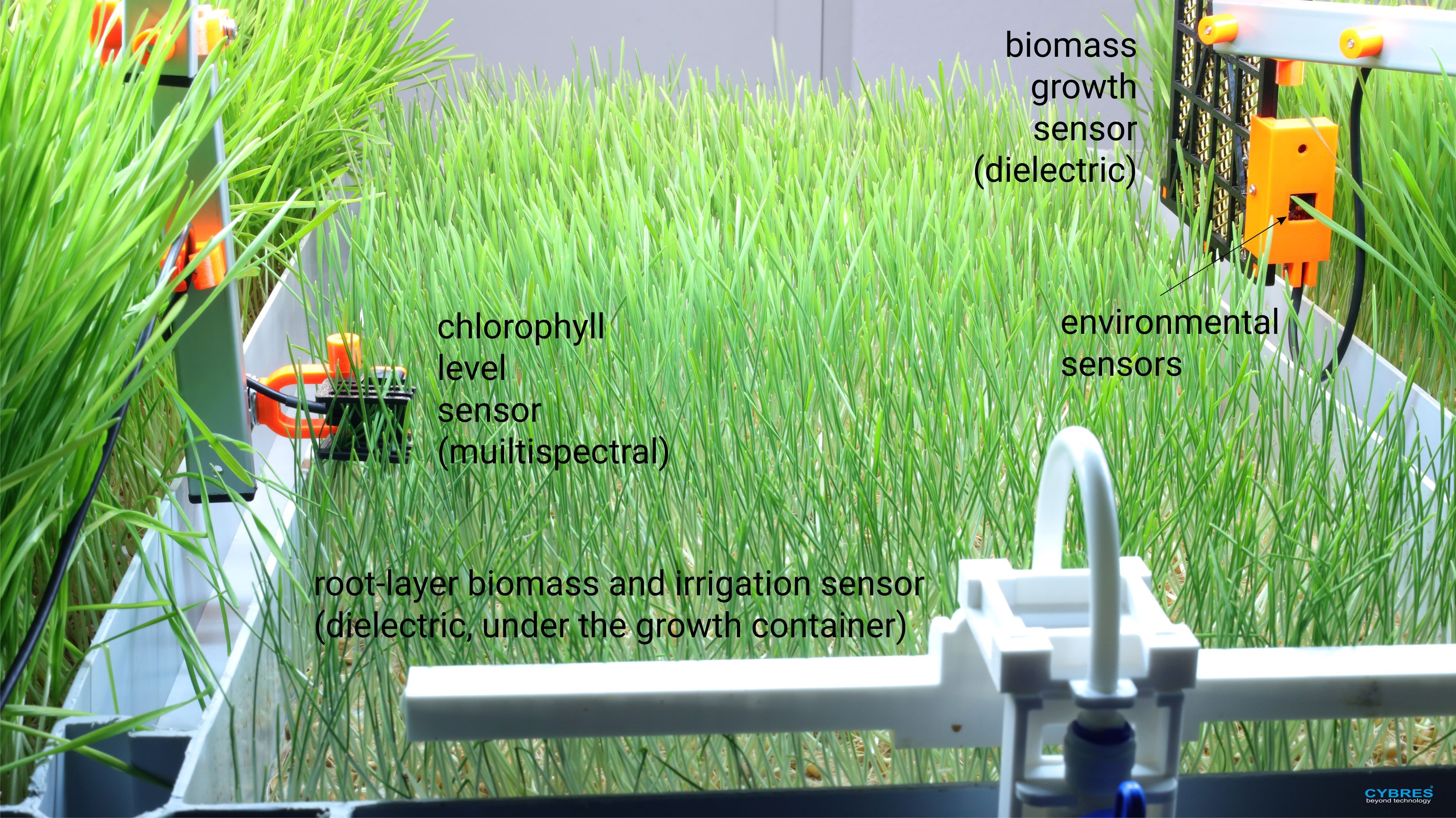}}
\caption{\small Experimental setups. \textbf{(a)} Multi-sensor framework with a \textit{Dracaena} plant; \textbf{(b)} High-density, soil-free vertical farming environment with \textit{Triticum aestivum} (wheatgrass) cultivation. \label{fig:setup}}
\end{figure}

For framework validation, \textit{Dracaena}, \textit{Solanum lycopersicum}, and \textit{Capsicum annuum} were deployed in parallel setups under controlled baseline conditions \cite{kernbach25EFS}, with \textit{Dracaena} serving as the primary case study (Fig. \ref{fig:setupA}). The closed-loop feedback system was further evaluated in a high-density, soil-free vertical farming environment cultivating wheatgrass (\textit{Triticum aestivum}) and green pea (\textit{Pisum sativum}) (Fig. \ref{fig:setupB}).

\begin{figure}[htp]
\centering
{\includegraphics[width=\columnwidth]{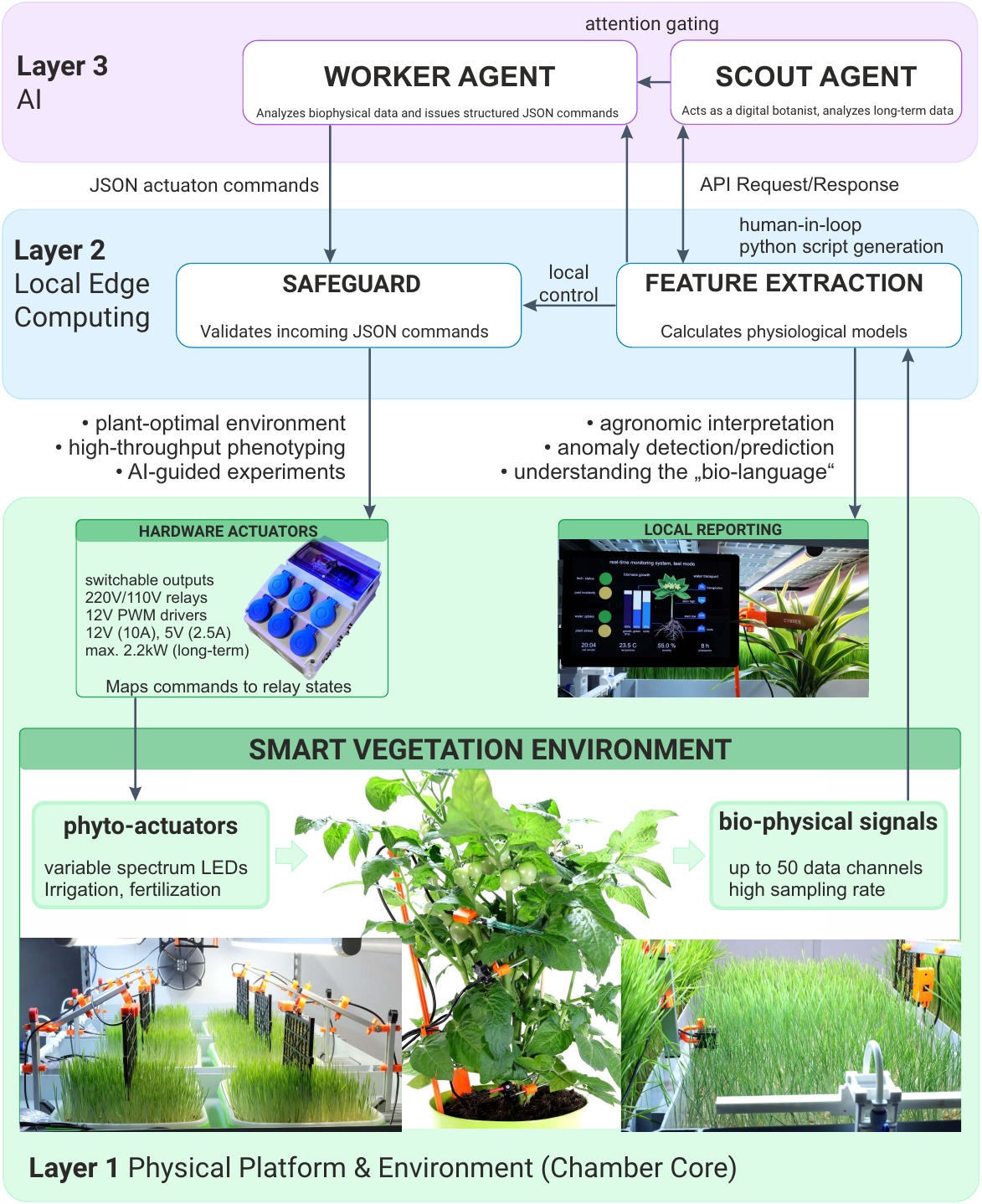}}
\caption{\small Structure of LLM driven closed-loop control.}
\label{fig:LLMdrivenControl}
\end{figure}

The multi-channel phytomonitoring network captures high-frequency raw data across electrochemical, dielectric, and optical domains \cite{CYBAPP28}. Table \ref{tab:channels} maps the functional interpretation of these raw readings and details how they are synthesized into secondary parameters acting as virtual sensors, providing the structured context delivered directly to the LLM.

Due to the morpho-anatomical constraints of high-density microgreen mats, the complete 49-channel sensor suite in the second setup was reduced to a specialized subset of sensors, see Table \ref{tab:closed_loop_column}:
\begin{itemize}
    \item \textbf{Dielectric Biomass Sensors:} Deployed within the canopy layer to log the real-time volumetric accumulation of fresh green biomass.
    \item \textbf{Root Zone Substrate Sensors:} Continuous monitoring of the root-mat matrix to capture water uptake dynamics, post-irrigation drainage profiles, net water retention, transpiration rate and water use efficiency.
    \item \textbf{Optical Multispectral Peripherals:} Sensor captures reflectance data at $515$, $555$, $690$, $745$, and $855$~nm to compute the three-band Vigor Index (VI) \cite{GITELSON1994286}, Photochemical Reflectance Index (PRI) and Normalized Difference Vegetation Index (NDVI) \cite{rouse1974monitoring,tucker1979red,gitelson1996use}.
\end{itemize}

\begin{table*}[htp]
\begin{center}
\caption{Prompt for AI: Fields in output data stream and analytical calculation formulas.} 
\label{tab:channels}
\fontsize{9}{11}\selectfont
\begin{tabular}{p{0.8cm} p{0.9cm} p{0.7cm} p{5.3cm} p{8.5cm}}
\hline \hline
\textbf{Field} & \textbf{Name} & \textbf{Units} & \textbf{Calculation} & \textbf{Description} \\ \hline

1 & time & -- & YY.MM.DD.HH.mm.ss & Time stamp of each measurement \\
2 & fr & Hz & $ch2 / 10$ & Frequency of the sweep for electrochemical (EIS) sensors \\ [3mm]

\multicolumn{5}{l}{\textbf{EIS-Channel 1: Electrochemical sensor (Lower stem area, close to roots, red label)}} \\ \hline
3 & $V_{I,max}$ & $\mu$V & -- & EIS: max. amplitude (upper peak) of sinus-excitation $V_I$ signal \\
4 & $V_{I,min}$ & $\mu$V & -- & EIS: min. amplitude (lower peak) of sinus-excitation $V_I$ signal\\
5 & $V_{lo}$ & Ohm & $ch5 / 1000$ & Impedance magnitude (based on RMS values), lower position \\
6 & phas & degree & $ch6 / 10000$ & Phase shift between $V_V$ and $V_I$ signals \\
7 & $V_{V,max}$ & $\mu$V & -- & EIS: max. amplitude (upper peak) of sinus-response $V_V$ signal  \\
8 & $V_{V,min}$ & $\mu$V & -- & EIS: min. amplitude (lower peak) of sinus-response $V_V$ signal  \\
9 & corr & -- & -- & Correlation between $V_V$ and $V_I$ signals for sweep frequency \\ [3mm]

\multicolumn{5}{l}{\textbf{EIS-Channel 2: Electrochemical sensor (Upper stem area, close to leaf, black label)}} \\ \hline
\multicolumn{5}{l}{10: $VI_{max}$ \quad 11: $VI_{min}$ \quad 12: $V_{up}$ Impedance magnitude, upper position \quad 13: phas \quad 14: $V_{V,max}$ \quad 15: $V_{V,min}$ \quad 16: corr} \\ [3mm]

\multicolumn{5}{l}{\textbf{Environmental Sensors}} \\ \hline
17 & $t_{PCB}$ & $^\circ$C & $ch17 / 10000$ & Temperature of electronic components in module \\
18 & $t_{ther}$ & $^\circ$C & $ch18 / 10000$ & Temperature of the thermostat \\
19--21 & mag & Gauss & $ch(19\text{--}21) / 1000$ & 3D magnetometer data (axes X, Y, Z), LIS3MDL sensor \\
22--24 & acc & g & $ch(22\text{--}24) / 1000$ & 3D accelerometer data (axes X, Y, Z), LIS3MDL sensor \\
25 & $t_{ext}$ & $^\circ$C & $ch25 / 10000$ & External temperature, LM35CA sensor \\
26 & light & lux & $\frac{ch26}{799.4} - 0.75056$ & Ambient light, APDS-9008-020 sensor \\
27 & rh & \% & $\frac{\frac{ch27 \times 3}{4200000} - 0.1515}{0.006707256 - 0.0000137376 \times \frac{ch25}{10000}}$ & External humidity, HIH-5031-001 sensor \\
30 & rf & -- & -- & RF power emission (450\,MHz--2.5\,GHz), MAX2204 sensor \\
33 & ap & mBar & -- & Air pressure, BMP280 sensor \\ [3mm]

\multicolumn{5}{l}{\textbf{Phytosensor Data}} \\ \hline
28 & dv1 & $\mu$V & -- & Differential bio-potential, channel 1 \\
29 & dv2 & $\mu$V & -- & Differential bio-potential, channel 2 \\
31 & tr & \% & $ch31 / 1000$ & Leaf transpiration sensor \\
32 & tSF & -- & -- & Thermal sap flow sensor (EIS spectrometer: coded fluid temperatures as XXXXXXYYYYYY, where XX.XXXX -- $t_{ch1}$, YY.YYYY -- $t_{ch2}$ in $^\circ$C) \\
34 & sm & \% & $\frac{ch34}{2175927} \times 100$ & Soil/rhizosphere moisture (root system water content) \\
35 & bm1 & \% & $\frac{ch35}{2175927} \times 100$ & Leaf biomass sensor (leaf water content) \\
36 & bm2 & \% & $\frac{ch36}{2175927} \times 100$ & Biomass sensor 2 (general tissue water content) \\
37--49 & opt & -- & $ch(37\text{--}49) / 65535$ & Multispectral 14-channel AS7343 sensor (405nm, 425, 450, 475, 515, 550, 555, 600, 640, 690, 745, 855, visual light) \\[3mm] 

\multicolumn{5}{l}{\textbf{Synthetic values, data after post processing}} \\ \hline
-- & $V_{lo\_z}$ & -- & $V_{lo\_z}=\frac{V_{lo}-\mu_{lo}}{\sigma_{lo}}$ & Global stDev (Z-score), mean ($\mu$) and standard deviation ($\sigma$) are calculated over the entire measurement period\\
-- & $V_{up\_z}$ & -- & $V_{up\_z}=\frac{V_{up}-\mu_{up}}{\sigma_{up}}$ & --  \\
-- & ${\Delta Z_t}$ & -- & ${\Delta Z} = \mu (V_{lo\_z, t}) - \mu(V_{up\_z, t})$ & Stem's hydrodynamic gradient between upper and lower EIS sensors, calculated daily or in sliding window of size $t$ \cite{10.1093/jxb/erq018} \\
-- & $V_{stem,t}$   & -- & $V_{stem,t}=|\Delta V_{t}-\Delta V_{t+1}|$, see$^1$  & Daily fluctuations of the hydrodynamic gradient compared to the previous day, RMS are calculated daily at $t$ and $t+1$ days \\
-- & $\text{CV}_{\text{Root}, t}$ & -- & $\text{CV}_{\text{Root}, t} = \frac{\sigma(V_{\text{lo}, t})}{\mu(V_{\text{lo}, t})}$ & Daily root zone CV (Coefficients of Variation): diurnal rhythms driven by root water uptake $\rightarrow$ $V_{\text{lo}}$ \\
-- & $\text{CV}_{\text{Canopy}, t}$ & -- & $\text{CV}_{\text{Canopy}, t} = \frac{\sigma(V_{\text{up}, t})}{\mu(V_{\text{up}, t})}$ & Daily canopy zone CV: diurnal rhythms driven by $V_{\text{up}}$ $\rightarrow$ transpiration\\
-- & $R_t$ & -- & $R_t=\frac{\sum_{i=1}^{N} (V_{\text{lo},i}-\bar{V}_{\text{lo}})(V_{\text{up},i}-\bar{V}_{\text{up}})}{\sqrt{\sum_{i=1}^{N} (V_{\text{lo},i}- \bar{V}_{\text{lo}})^2 \sum_{i=1}^{N} (V_{\text{up},i} - \bar{V}_{\text{up}})^2}}$ & Daily root-canopy correlation, see$^2$: $R$ indicates a unified hydraulic conduit from root water uptake to  canopy transpiration -- coordinated ionic and mass-flow flux throughout the stem\\
-- & PRI & -- & $\text{PRI} = \frac{R_{515} - R_{555}}{R_{515} + R_{555}}$ & Photochemical Reflectance Index: Light-use efficiency tracking linked to the xanthophyll pigment cycle kinetics, e.g. $R_{515}$ is 515nm channel \cite{GAMON199235}\\
-- & NDVI & -- & $\text{NDVI} = \frac{R_{855} - R_{690}}{R_{855} + R_{690}}$ & Normalized Difference Vegetation Index (NDVI) quantifies how strongly chlorophyll absorbs red light for photosynthesis and how efficiently leaf structure reflects near-infrared radiation \cite{gitelson1996use}\\ 
-- & VI & -- & $\text{VI} = \left( \frac{1}{R_{690}} - \frac{1}{R_{745}} \right) \times R_{855}$ & Vigor Index by Gitelson \cite{GITELSON1994286}: three-band parameter for total canopy chlorophyll content estimation \\\hline
\end{tabular}
\end{center}
$^1$ $\Delta V_{t}=\text{RMS}(V_{lo\_z,t})-\text{RMS}(V_{up\_z,t})$, $\Delta V_{t+1}=\text{RMS}(V_{lo\_z,t+1})-\text{RMS}(V_{up\_z, t+1})$
\newline $^2$ $N$ is the total number of samples within day $t$, $V_{\text{lo}, i}$ and $V_{\text{up}, i}$ are impedance measurements at timestamp $i$, $\bar{V}_{\text{lo}}$ and $\bar{V}_{\text{up}}$ are the daily means for day $t$.
\end{table*}

The closed-loop LLM-driven control architecture is structured as an edge-agent hybrid framework (see Fig. \ref{fig:LLMdrivenControl}). Low-level phyto-actuation \cite{CYBAPP28} and high-frequency data ingestion are managed by a local Python script. Exploration algorithms, state-space evaluation, and LLM orchestration reside within the AI layer, which outputs a predictive control JSON sequence executed autonomously by the edge hardware. To prevent context dilution within the LLM, the AI layer deploys a Worker Agent (WA) and a Scout Agent (SA). The SA acts as an attention router: it parses long-term history to isolate systemic anomalies and rewrites prompt for WA. The WA then processes real-time data, main and focusing prompts to synthesize the final control sequences. In conducted tests, the WA and SA were evaluated using both proprietary cloud-based models and local open-source architectures.

To mitigate LLM hallucinations \cite{10.1145/3703155}, the framework implemented a three-tier verification approach:
\begin{enumerate}
    \item \textbf{Contextual Grounding:} Restricts LLM reasoning strictly within the real-time sensor matrices (Table \ref{tab:channels}).
    \item \textbf{Syntactic Constraints:} Restricts the model's output to a rigid, predefined JSON schema.
    \item \textbf{Operational Verification:} The local Python script outputs console-based control tables from raw sensor data; these logs are fed back into the LLM context for real-time validation and sanity checks.
\end{enumerate}

\section{Case Study 1: LLM driven closed-loop control}

\begin{figure}[htp]
\centering
\subfigure{\includegraphics[width=\columnwidth]{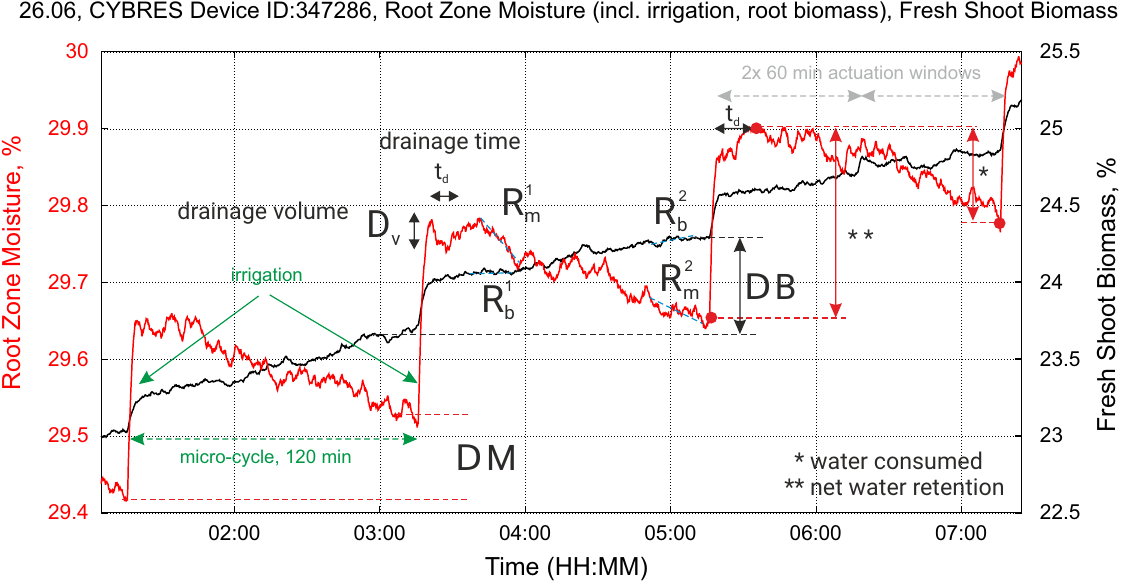}}
\caption{\small Dynamics of root zone moisture (with irrigation and root biomass growth effects) and fresh shoot biomass during consecutive micro-cycles.}
\label{fig:structureCase3_microcycle}
\end{figure}

In the first case study, the LLM operates as a real-time, closed-loop biofeedback controller. Thin-film NFT cultivation in $40 \times 80$ cm containers inside a vertical farm, see Fig. \ref{fig:setupB}, is divided into 120-minute irrigation micro-cycles. Each micro-cycle contains 10-to-60-minute actuation windows, where the LLM switches light and other phyto-actuators. Dynamics of root zone moisture (RZM) and fresh shoot biomass (FSB) are shown in Fig. \ref{fig:structureCase3_microcycle} and are used to calculate several secondary parameters. Following a 20-minute irrigation drainage phase, the system estimates the biomass rate ($R_b$), moisture rate ($R_m$), drainage volume $D_v$, net water retention of the root layer and consumed water. Additional tracked parameters include RZM and FSB increase between micro-cycles as $\Delta M$ and $\Delta B$, water use efficiency (WUE) and photoperiod latency, see Table \ref{tab:closed_loop_column}. $R_b$ and $R_m$ are calculated for each actuation window, while the remaining parameters are computed for the full micro-cycle. This architecture demonstrates how two raw sensor channels yield 8 to 12 derived parameters per evaluation cycle.

\begin{table}[htp]
\centering
\fontfamily{ptm}\selectfont
\footnotesize 
\caption{Closed-Loop Biofeedback Configuration Matrix.}
\label{tab:closed_loop_column}
\begin{tabularx}{\linewidth}{>{\raggedright\arraybackslash}p{2.0cm} >{\raggedright\arraybackslash}X}
\toprule
\textbf{Component} & \textbf{Operational Target \& LLM Control Logic} \\
\midrule
\multicolumn{2}{l}{\textbf{Sensors (System Inputs)}} \\
\addlinespace
Canopy biomass & \textbf{Fresh shoot biomass ($\text{FSB}$) \& turgor.} Primary feedback (\textit{bm2, ch.36}). \\
Root biomass and irrigation  & \textbf{Root zone moisture (RZM).} Analyzed by LLM to compute root level parameters (\textit{sm, ch.34}). \\
Spectral Sensor & \textbf{Multispectral optical feedback} \textit{ch.37-49}. \\
Environment   & temperature, light, air humidity (\textit{$t_{ext}$ ch.25, light ch.26, rh ch.27}) \\
\midrule
\multicolumn{2}{l}{\textbf{Computed Secondary Parameters (Virtual Sensors)}} \\
\addlinespace
Net Water Retention, $\Delta M$ & \textbf{Substrate moisture capacity and root growth} between micro-cycle N and N-1, delta between post-drainage stabilization and the pre-irrigation baseline 
$\text{RZM}[N]_{\text{stable}} - \text{RZM}[N-1]_{\text{stable}}$\\
Delta Biomass, $\Delta B$ & \textbf{Fresh shoot biomass growth} between micro-cycles, $\text{FSB}[N]_{\text{stable}} - \text{FSB}[N-1]_{\text{stable}}$. \\
Drainage Volume, $D_v$ & \textbf{Gravitational runoff dynamics}, the sharp decay following the saturation peak $\text{RZM}_{\text{peak}}-\text{RZM}_{\text{stable}}$ to optimize total irrigation efficiency. \\
Transpiration Rate, $R_M$ & \textbf{Stomatal activity dynamics}, slope of post-drainage substrate moisture decay to monitor the transpirational pull. \\
Biomass Rate, $R_B$ & \textbf{FSB growth dynamics}, slope of post-drainage biomass growth to monitor the biomass response to light and irrigation. \\
Water Use Efficiency, WUE & \textbf{Metabolic cost optimization}, the ratio of $\Delta B$ to $\Delta M$; used by LLM to balance growth versus resource depletion. \\
Photoperiod Latency & \textbf{Stomatal kinetics assessment}, the temporal delay ($\Delta t$) between full-spectrum activation and $R_B$, $R_M$ dynamics to assess canopy health. \\
\midrule
\multicolumn{2}{l}{\textbf{Phyto-actuators (System Outputs)}} \\
Full Spectrum, \newline \textit{command: 'wl0' OFF, 'wl1' ON}    & \textbf{Baseline PAR solar simulation.} Sustains background photosynthesis; ensures deep foliar canopy penetration; 37.5 $W/m^2$  \\
Blue LED (450 nm), \textit{command: 'wi0/1'} & \textbf{Cryptochrome \& stomatal kinetics.} Triggers stomatal opening to actively drive transpirational pull and sap mass flow; 6.25 $W/m^2$ \\
Red LED (660 nm), \textit{command: 'wq0/1'} & \textbf{Chlorophyll $a/b$ excitation.} Main engine; dynamically boosted by LLM to accelerate baseline biomass accumulation; 12.5 $W/m^2$ and 39 $W/m^2$ \\
Far-Red (730-745nm); \newline UV 365-420nm
& \textbf{Phytochrome equilibrium}, to stimulate leaf blade expansion if canopy growth stalls; \textbf{Stress hardening.} Applied in short pulses for mechanical tissue reinforcement and pathogen suppression. \\
Irrigation Unit & \textbf{Soil-free hydration}, 2h intervals with basic NPK. \\
Stimulation Unit & \textbf{Non-chemical stimulation} with EM fields.\\
Actuation micro-cycle & Executes discrete control steps (10--60 min range), dynamically re-evaluated every 120 minutes. \\
Growth macro-cycle & Spans the full crop production cycle, which varies between 4 and 7 days for wheatgrass. \\
\midrule
\multicolumn{2}{l}{\textbf{Objective Functions}} \\
Time-Optimal & $\max \frac{d(\text{FSB})}{dt}$. Maximizes biomass accumulation rate over time to achieve the fastest crop development. \\
Energy-Efficient & $\max \frac{\Delta \text{FSB}}{Wh}$. Maximizes $\Delta B$ per unit of consumed electrical energy, minimizing operational costs. \\
\bottomrule
\end{tabularx}
\end{table}

From a physiological perspective, $R_b$, $R_m$, $D_v$, $\Delta M$, and $\Delta B$ serve as a real-time proxy matrix for a wide range of physiological parameters, including hydrodynamic transpiration, irrigation efficiency, root anoxia, and overall plant vitality. The actuation window duration is determined by the kinetics of stomatal opening; it must be sufficient to register stabilization in stomatal conductance and accurately calculate $R_b$ and $R_m$. Stomatal dynamics initiate within 2 to 5 minutes and reach a steady state within 30 to 60 minutes \cite{10.1146/annurev.arplant.57.032905.105434, bg-21-1501-2024}. Evaluation of 10-, 30-, and 60-minute intervals demonstrate that, due to stomatal and hydrodynamic inertia, a 60-minute actuation window is optimal.

The LLM utilizes the derived hydrodynamic and multispectral time-series to optimize crop growth via light and irrigation actuators. Although dynamic spectral manipulation is established in precision agriculture \cite{jensen2025dynamic, monostori2018led}, optimization must account for complex physiological codependencies. Fresh biomass accumulation occurs predominantly during the dark phase. During the photoperiod, photosynthesis drives carbohydrate synthesis, while high transpiration rates limit physical cell expansion. In the dark phase, suppressed transpiration elevates root-driven turgor pressure, enabling cell elongation and measurable biomass increase driven by accumulated sugars and auxins. To maximize this accumulation, the controller alternates photoperiods -- combining baseline full-spectrum light with supplemental red and blue channels to maximize carbohydrate synthesis -- with dark phases optimized for turgor-driven elongation. Simultaneously, active apical wheat roots consume up to 3.8~nmol~O$_{2}$~g$^{-1}$~\textit{FM}~s$^{-1}$, creating a hypoxia risk in thin-film hydroponics \cite{10.1093/jxb/erag030, 10.1071/PP9880599}. To prevent root anoxia during turgor-driven dark periods, the controller must co-optimize irrigation frequency, forced aeration, and spectral distribution to align actuation with cyclic metabolic demands.

Conventional control methods are limited in this task because plant physiology and nutrient chemistry are non-linear and lack explicit mathematical models, requiring the integration of agronomic and biological expert knowledge. This reliance on non-formalized reasoning justifies the application of an LLM capable of qualitative context analysis. However, processing this domain knowledge is complex: three raw sensor channels generate 15 to 18 distinct parameters, and the system must simultaneously perform high-level physiological analysis and low-level phyto-actuator control. This functional task divergence causes LLM context dilution. To resolve this, the system implements a two-agent topology operating at different timescales. The scout agent (SA) runs once daily to analyze historical telemetry and identify physiological or environmental anomalies (detailed in Case Studies 2 and 3). The worker agent (WA) executes per micro-cycle to process time-series data and synthesize control sequences based on agronomic rules, focusing on the anomalies flagged by the SA. Approximately 30\% of the WA prompt capacity is allocated to inputs from the SA.

This study evaluates two objective functions: time-optimal and energy-optimal control. The former maximizes the shoot biomass growth rate, formulated as $\max J_{\text{time}} = \frac{d(\text{FSB})}{dt}$, while the latter maximizes biomass energy efficiency: $\max J_{\text{energy}} = \frac{\Delta \text{FSB}}{Wh}$. For comparison, the optimized scenarios are evaluated against a periodic baseline control. This baseline uses a 12/12 photoperiod for full-spectrum lighting (37.5~$\text{W/m}^2$) with constantly active supplementary red LEDs (24~h, 6.25~$\text{W/m}^2$), representing the minimum threshold for wheat cultivation \cite{SLAFER19961}.

\begin{figure}[htp]
\centering
\subfigure{\subfigure{\includegraphics[width=\columnwidth]{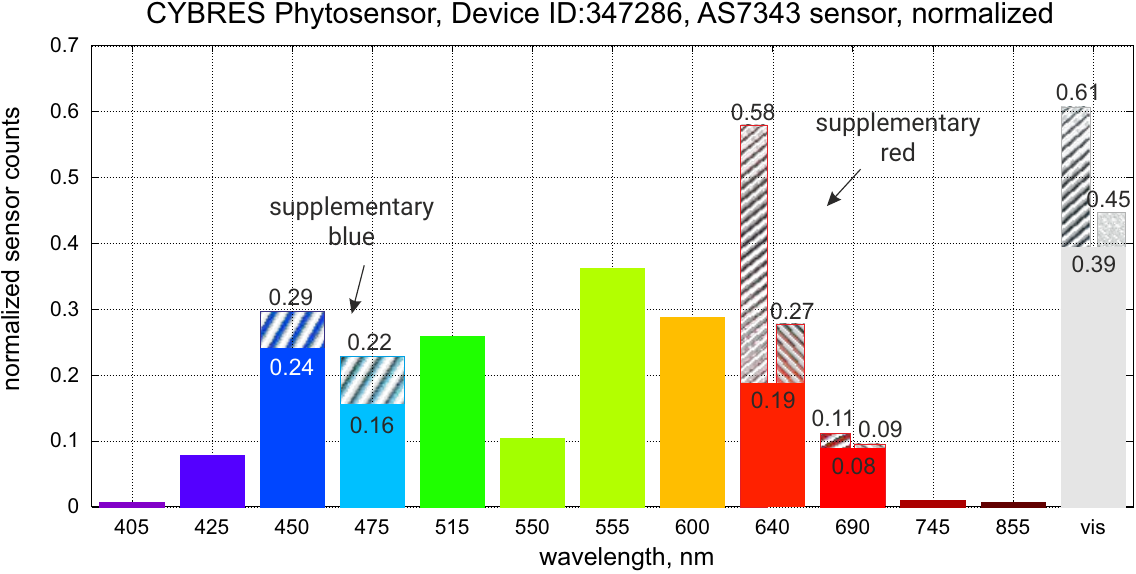}}}
\caption{\small Full spectrum light with supplementary blue (450-475nm) and two versions of supplementary red (640-690nm) light.}
\label{fig:supplementraSpectrum}
\end{figure}

Under these optimization criteria, the worker agent generates command sequences that govern the full-spectrum and narrow-band (450~nm and 660~nm) lighting channels (Table \ref{tab:closed_loop_column}). Two configurations of supplementary red lighting were initially considered, providing spectral irradiance enhancements of +205\% and +42\% at 640~nm (corresponding to 12.5~$\text{W/m}^2$ and 39~$\text{W/m}^2$, respectively; see Fig. \ref{fig:supplementraSpectrum}). To avoid overheating, the low-power red LED variant was deployed. The supplementary blue lighting provides an enhancement of +20.8\% at 450~nm and +37.5\% at 470~nm with 6.25~$\text{W/m}^2$. Looking ahead, Far-Red (730-745~nm) and UV (365-420~nm) bands, along with irrigation and growth stimulation units can be integrated as supplementary actuators in further closed-loop control scenarios.

\begin{table}[htp]
\centering
\fontfamily{ptm}\selectfont
\footnotesize 
\caption{LLM Control Algorithms and Operational Modes.}
\label{tab:control_algorithms}
\begin{tabularx}{\linewidth}{>{\raggedright\arraybackslash}p{1.7cm} >{\raggedright\arraybackslash}X}
\toprule
\textbf{Operational Mode} & \textbf{LLM Execution Logic \& Edge Action} \\
\midrule
Exploration (8--12 actuation steps) & \textbf{Heuristic Mutation:} The LLM rationalizes decisions using genetic algorithm principles, acting textually as a mutation operator over historical logs to propose candidate light matrices. \\
\midrule
Exploration (2--4 actuation steps in micro-cycle) & \textbf{Text-Based Adaptation \& Reinforcement Learning Logic:} The agent structures its reasoning around an $\epsilon$-greedy strategy, evaluating $\frac{d(\text{FSB})}{dt}$ as a metabolic reward to select lighting profiles. \\
\midrule
Predictive Exploitation & \textbf{Model Predictive Control Reasoning:} Triggered during growth plateaus or stress signals. The LLM textually projects system states onto multi-hour horizons to guide stabilization. \\
\midrule
Local Edge Execution & \textbf{Deterministic Action Sequence:} Stores generated daily actuation sequences; functions independently after 10–15 macro-cycles.\\
\bottomrule
\end{tabularx}
\end{table}

Based on physiological feedback, the worker agent alternates between exploration and exploitation modes (Table~\ref{tab:control_algorithms}). Every micro-cycle, the model processes historical and current telemetry to generate lighting configurations formatted as a JSON matrix for the local edge controller. This framework stores actuation sequences during growth macro-cycles. To minimize AI usage under identical cultivar and environmental conditions, the local edge can operate autonomously using cached sequences, escalating to the worker agent only when the scout agent detects anomalies. If both agents are unavailable, the edge operates as a deterministic state-machine using cached or default actuation sequences.

The optimization prompt for WA is structured around objective functions and the number of actuation windows within a micro-cycle (Table~\ref{tab:control_algorithms}). For 10-minute windows (12 actuation steps), the agent's logic relies on combinatorial analysis and heuristic mutation operators to synthesize spectral configurations from historical logs. For fewer actuation steps, the model textually applies an $\epsilon$-greedy reinforcement learning strategy to select lighting profiles based on the biomass growth reward. To account for non-linear physiological responses, the prompt guides the agent toward model predictive control principles, projecting plant states over multi-hour horizons. The SA references biological and agronomic models to detect physiological anomalies. Through cross-model correlation analysis of multichannel time-series data, the agent generates directives that constrain the search space within the WA instructions (detailed in Case Studies 2 and 3). Both agents were validated across cloud-based and local LLM architectures, including proprietary and open-weight models.

\begin{figure}[htp]
\centering
\subfigure[\label{fig:heatMapControl}]{\includegraphics[width=\columnwidth]{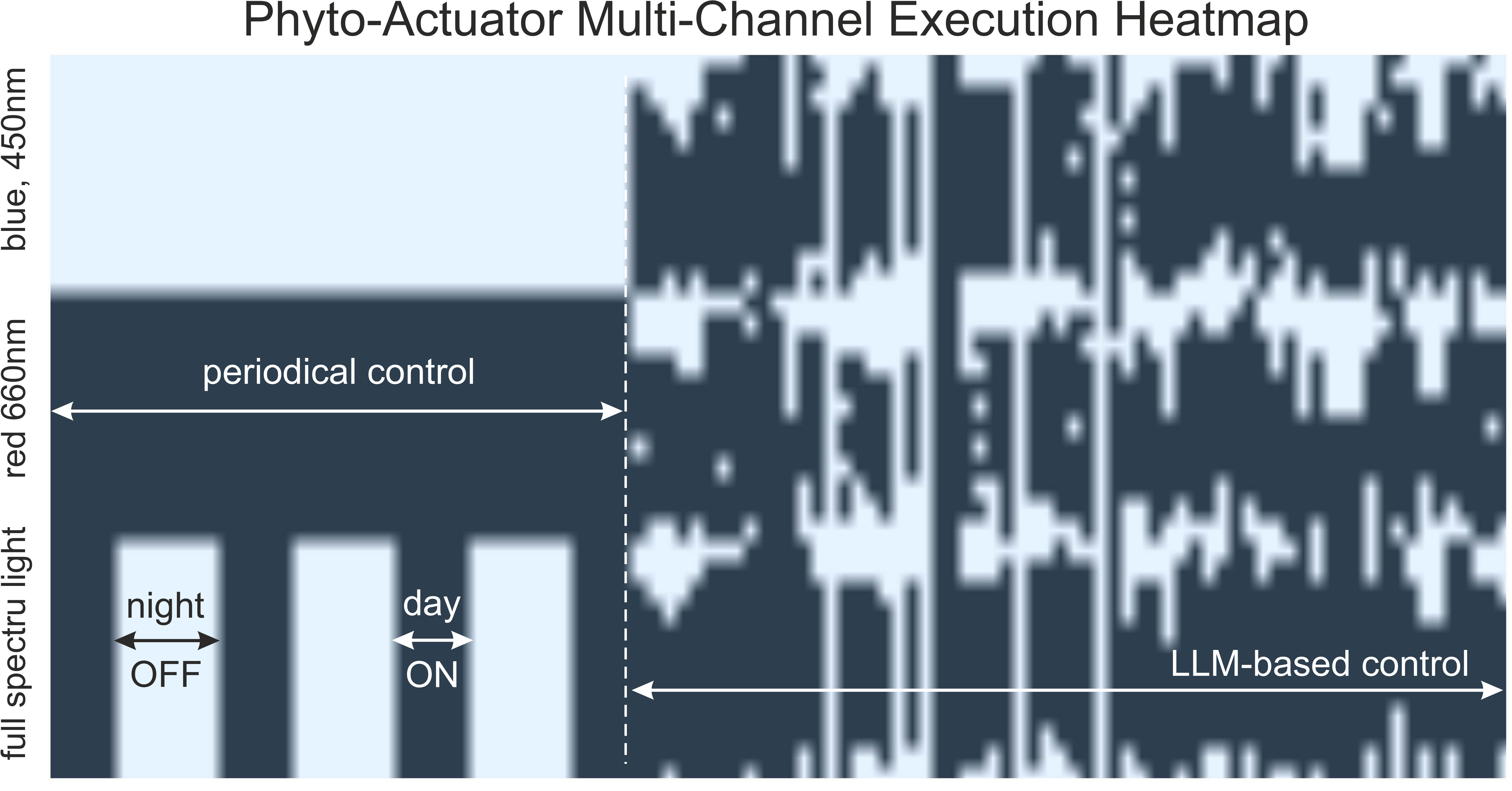}}
\subfigure[\label{fig:wheatgrassHeight}]{\includegraphics[width=\columnwidth]{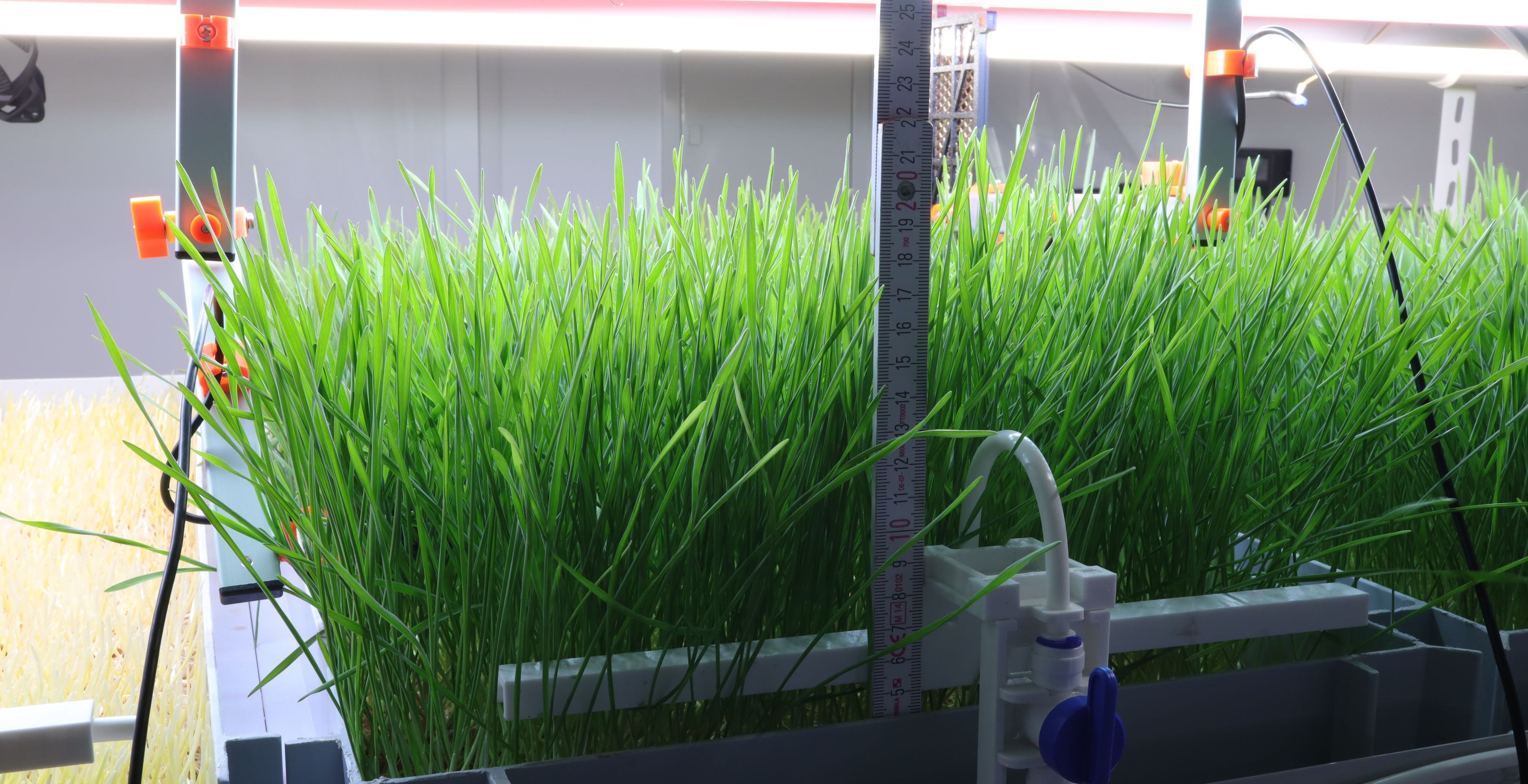}}
\caption{\small \textbf{(a)} Example of phyto-actuator execution heatmap under periodical (10/14 photoperiod) versus LLM-driven control (10 min. actuation step); \textbf{(b)} Visual verification of the wheatgrass height (19–20 cm) achieved within a 4.5-day light-induced growth phase under LLM-driven control (minimal time).}
\end{figure}

\begin{figure}[htp]
\centering
\subfigure[\label{fig:comparison}]{\includegraphics[width=\columnwidth]{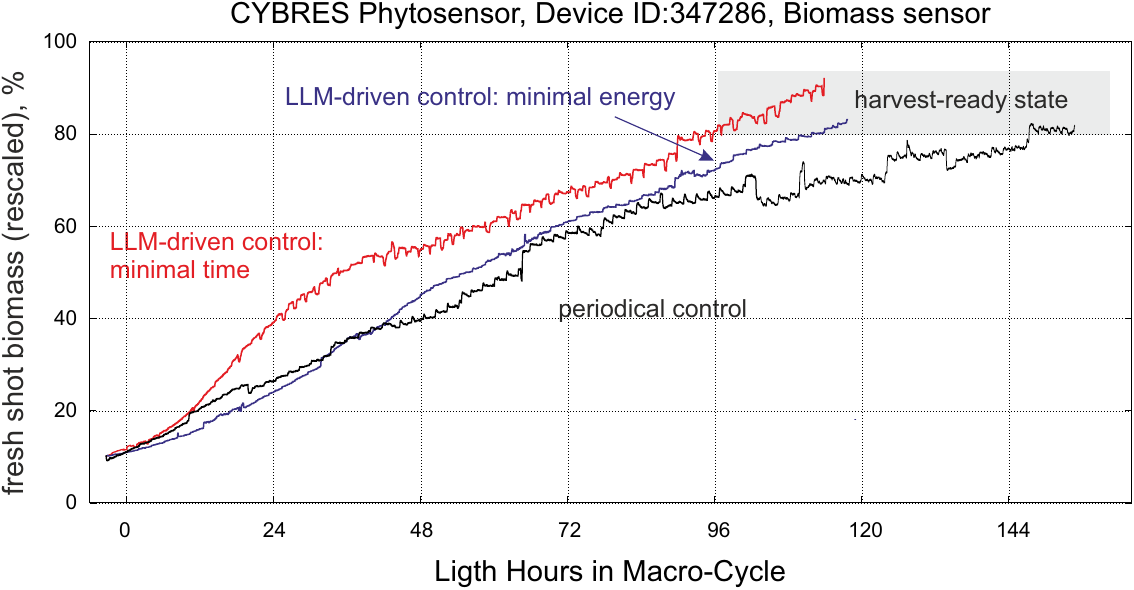}}
\subfigure[\label{fig:comparisonVI}]{\includegraphics[width=\columnwidth]{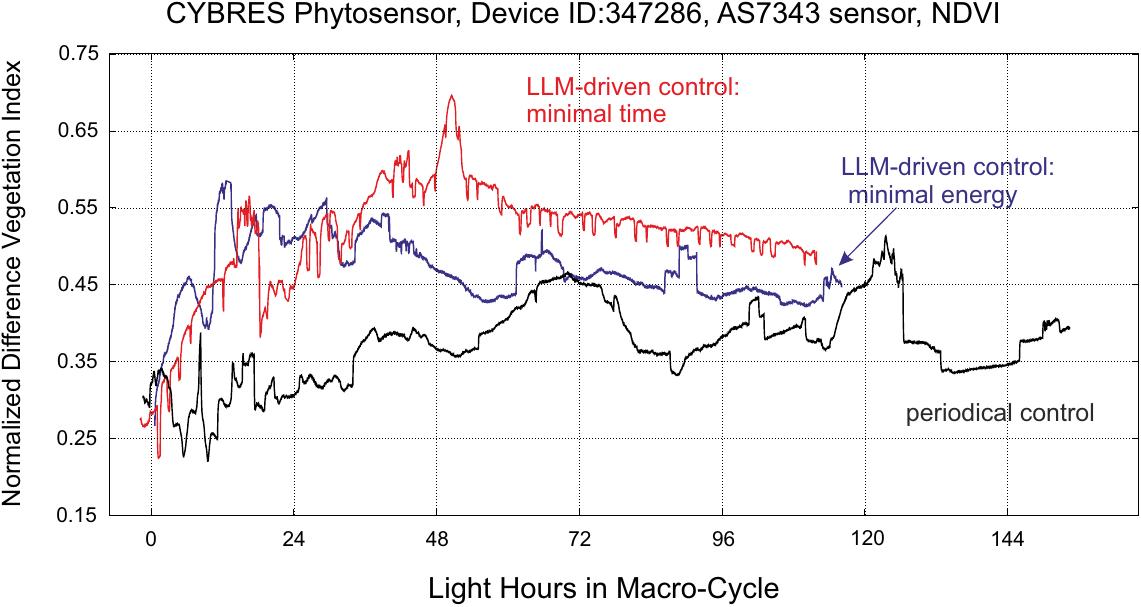}}
\caption{\small Example of macro-cycle vegetation monitoring under periodical and LLM-driven control: \textbf{(a)} Fresh shot biomass accumulation kinetics within the first 48 hours of cultivation. The relative biomass values are rescaled between 0 and 100\% (where 80\%-100\% corresponds to the wheatgrass harvest-ready state) across all experimental runs, clearly demonstrating linear (periodical control), sublinear (minimal energy control), and superlinear (minimal time control) growth dynamics during first 48 hours. \textbf{(b)} Temporal NDVI profiles. While the supplementary spectral light alters the sensor's absolute calibration baseline and introduces measurement artifacts, the relative behavioral trends and comparative performance between all control loops remain consistent. \label{fig:comparison2}
}
\end{figure}

\textbf{Minimal Time Optimization.} A comparative analysis of periodic (diurnal) and LLM-driven control is illustrated in Figs. \ref{fig:heatMapControl} and \ref{fig:comparison}. Following the exploration phase, the agent extended the photoperiod to 22/2 or 24/0 hours, consistent with speed breeding methodologies \cite{watson2018speed, bugbee1988exploring, ruzvidzo2024light}. While applicable to wheat as a long-day crop, photoperiod-sensitive species like tomato would require different actuation constraints, generated dynamically due to risks of chlorosis and physiological injury. For wheat, the LLM modulated supplemental spectral channels to optimize the daily light integral while mitigating photoprotective stress and photosynthetic saturation. Periodic control resulted in linear biomass growth and a monotonic increase in NDVI over 144 hours, see Fig. \ref{fig:comparison2}. Conversely, the minimal time mode induced superlinear biomass kinetics by optimizing micro-climatic inputs for shoot elongation, resulting in an NDVI peak at 48 hours before stabilizing. Under light-induced growth, the wheatgrass achieved a height of 19--20 cm within 4.5 days (Fig. \ref{fig:wheatgrassHeight}), permitting juice extraction by day 4.

\textbf{Minimal Energy Optimization.} Although mature wheat requires a baseline photoperiod of 12 hours \cite{SLAFER19961}, younger wheatgrass remains viable under shorter light cycles. Furthermore, a mild light deficit induces seedling elongation while suppressing cell wall lignification; this reduces tissue rigidity and enhances leaf succulence, maximizing wheatgrass juice extraction yield \cite{Kulkarni06}. A critical risk of this strategy is optimization convergence toward a 0/24 photoperiod, where the agent deactivates the LEDs to eliminate energy costs ($Wh$). To prevent this, the system maps the objective function numerator ($\Delta \text{FSB}$) directly to the shoot biomass sensor. If biomass accumulation stagnates ($\Delta \text{FSB} \to 0$), the efficiency ratio drops to zero regardless of energy savings, forcing the LLM to activate the illumination. Moreover, the agent minimizes the denominator ($Wh$) by shifting from continuous full-spectrum lighting ($37.5\,\text{W/m}^2$) to intermittent spectral configurations, alternating low-power blue ($6.25\,\text{W/m}^2$) and red ($12.5\,\text{W/m}^2$) channels within 10- to 60-minute intervals.

\begin{figure}[htp]
\centering
\subfigure{\includegraphics[width=\columnwidth]{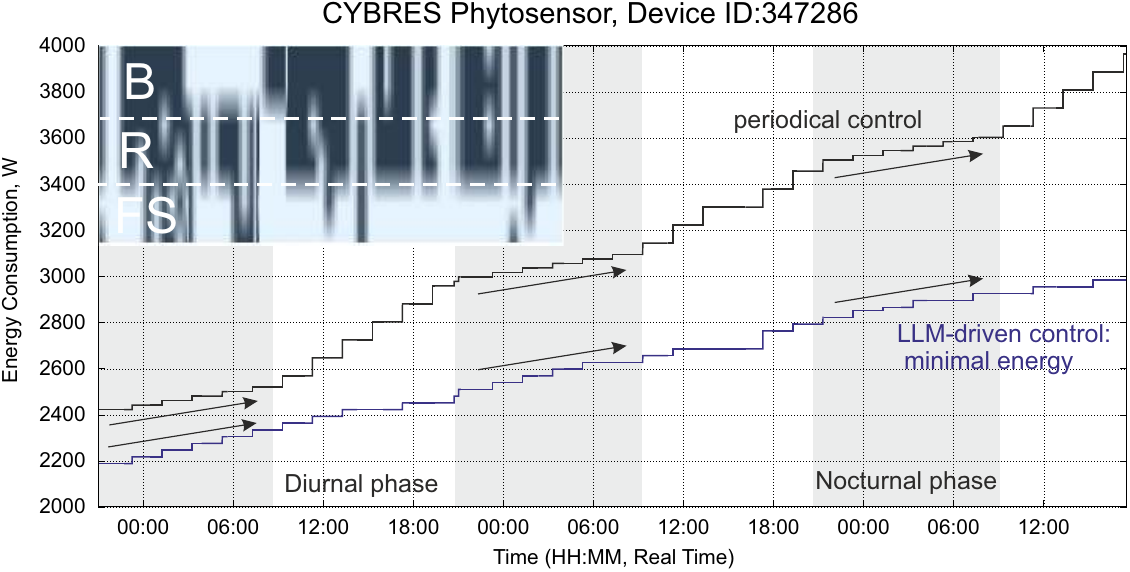}}
\caption{\small Energy-saving control mechanisms during the final 70 hours of the growth macro-cycle under a 120-minute actuation micro-cycle. The system reduces full-spectrum lighting and introduces intermittent dark phases decoupled from standard diurnal rhythms to optimize endogenous crop kinetics and to minimize energy consumption.}
\label{fig:dark}
\end{figure}

The controller adjusted light-dark intervals based on the photosynthetic induction and physiological inertia of the plants \cite{kaiser2015dynamic, kanechi2016effects, jishi2012pulsed}. By timing light periods to align with the kinetics of the photochemical apparatus, the system reduced dark-phase energy consumption without reducing biomass accumulation. While conventional studies utilize high-frequency lighting (typically 40–50 Hz) to optimize electron transport, the agent implemented a low-frequency, intermittent regime. This strategy operates on a longer timescale, utilizing stomatal inertia and physiological relaxation periods to reduce energy consumption without inducing protective stomatal closure. Both energy conservation mechanisms are demonstrated in Fig. \ref{fig:dark}. Rather than maintaining standard 12-hour diurnal phases, the system stabilizes the crop under an endogenous rhythm optimized for wheat physiology. As shown in Fig. \ref{fig:comparison2}, the minimal energy mode induces a sublinear biomass trend due to resource conservation and extends the duration of biomass development to 5 days, representing an 11\% to 25\% increase in cultivation time. 

A comparison of the time and energy metrics across the three operational modes is summarized in Table \ref{tab:energy_comparison}, based on five growth macro-cycles for each configuration. The minimum-time strategy shortened the cultivation cycle by 2.25 days relative to the periodic control but required 48\% more total energy. Conversely, the minimum-energy mode extended the cycle duration to 5.0 days compared to the minimum-time strategy, yet achieved an 18\% energy reduction relative to the periodic baseline and a 45\% reduction compared to the minimum-time strategy. 

\begin{table}[htp]
\centering
\fontfamily{ptm}\selectfont
\footnotesize 
\caption{Duration and Energy Consumption per $m^2$ per growth cycle.}
\label{tab:energy_comparison}
\begin{tabularx}{\linewidth}{>{\raggedright\arraybackslash}p{1.0cm} >{\centering\arraybackslash}X >{\centering\arraybackslash}X >{\centering\arraybackslash}X >{\centering\arraybackslash}X >{\centering\arraybackslash}X}
\toprule
\textbf{Scenario} & $\mathbf{t_{\text{growth}}}$ \newline (days) & $\mathbf{kWh_{\text{FS}}}$ & $\mathbf{kWh_{\text{R}}}$ & $\mathbf{kWh_{\text{B}}}$ & \textbf{Total} \newline $\mathbf{kWh}$ \\
\midrule
\textbf{Periodic}   & 6.5$\pm$0.5 & 2.28$\pm$0.2 & 0.76$\pm$0.1 & 0   & 3.04$\pm$0.2  \\
\addlinespace
\textbf{Min Time}   & 4.25$\pm$0.25 & 3.13$\pm$0.1 & 0.89$\pm$0.02 & 0.48$\pm$0.01 & 4.49$\pm$0.01 \\
\addlinespace
\textbf{Min Energy} & 5.0$\pm$0.25 & 1.05$\pm$0.05 & 1.07$\pm$0.05  & 0.37$\pm$0.02 & 2.49$\pm$0.12 \\
\addlinespace
\textbf{Ultra Min} & 5.25$\pm$0.25 & 0              & 0.52$\pm$0.03  & 0.28$\pm$0.01 & 0.8$\pm$0.04 \\
\bottomrule
\end{tabularx}
\end{table}

\begin{figure}[htp]
\centering
\subfigure[\label{fig:chlorophyll1}]{\includegraphics[width=\columnwidth]{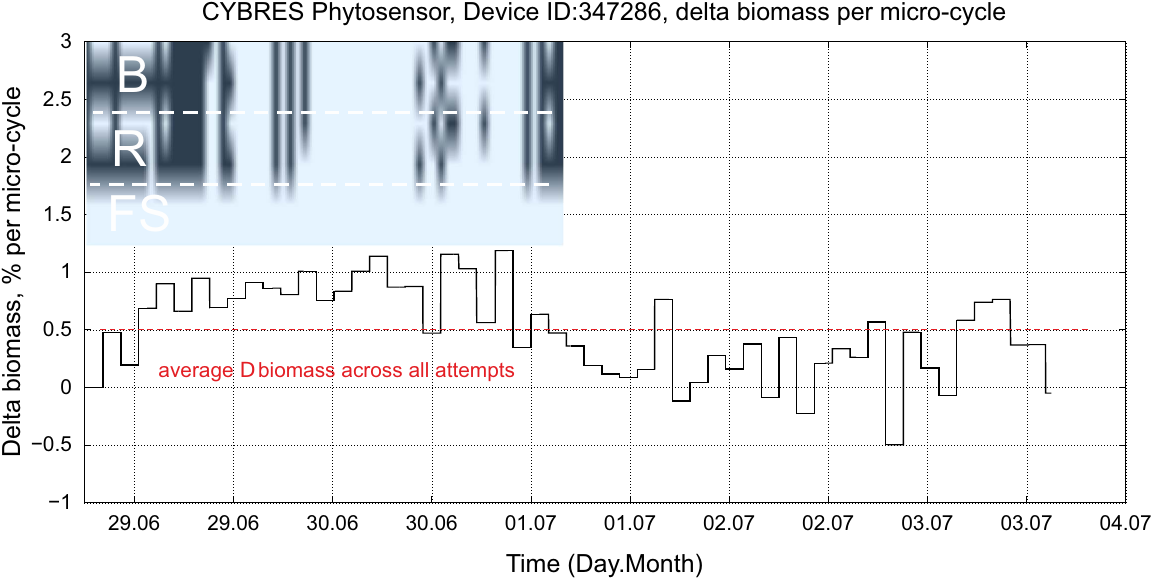}}
\subfigure[\label{fig:chlorophyll2}]{\includegraphics[width=\columnwidth]{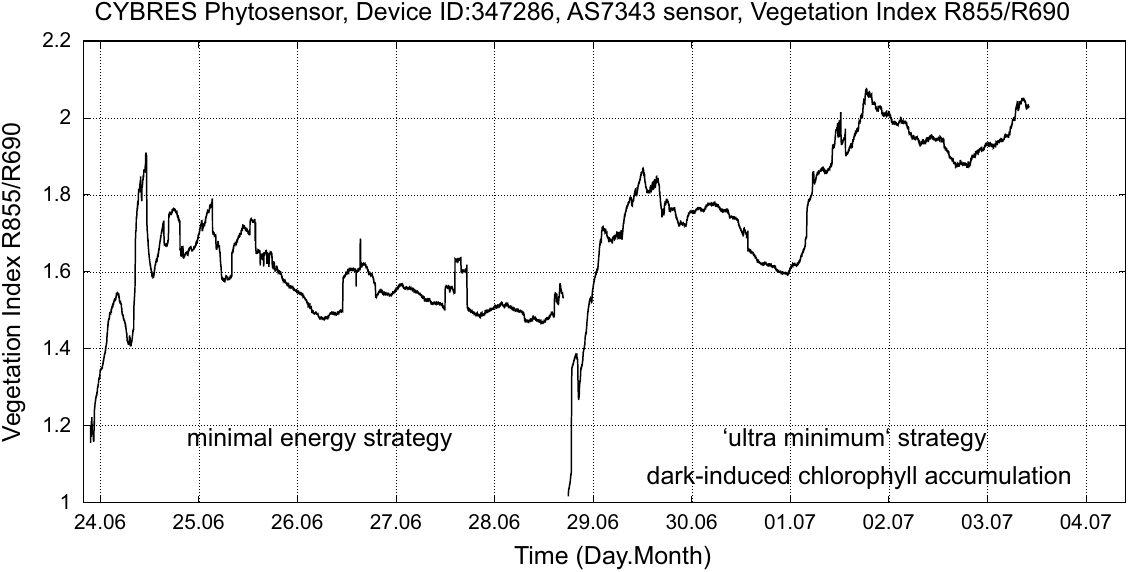}}
\subfigure[\label{fig:chlorophyll3}]{\includegraphics[width=\columnwidth]{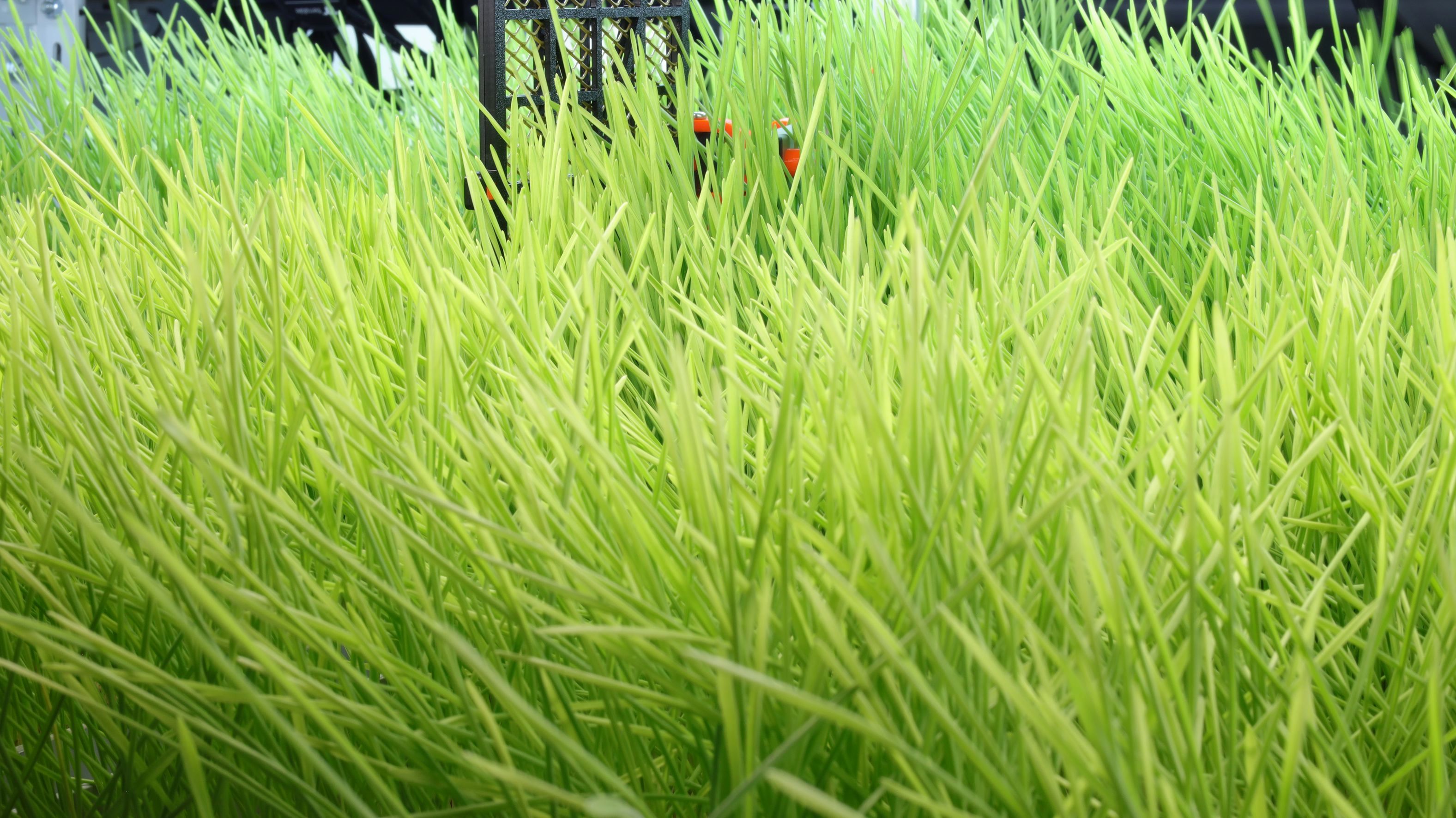}}
\caption{\small Emergent 'Ultra Minimum' strategy of dark-induced chlorophyll accumulation developed by agents: \textbf{(a)} Examples of biomass growth dynamics per micro-cycle and actuation heatmap; \textbf{(b)} Vegetation Index $R_{855}/R_{690}$ comparing two successive runs under minimal energy and 'ultra minimum' strategies. The spectral light alters the sensor's absolute calibration and introduces measurement artifacts, the comparative performance between all two attempts  remain consistent; \textbf{(c)} Visual observation of wheatgrass exhibiting a light-green hue. \label{fig:chlorophyll}}
\end{figure}

Observations show that the AI maps and explores biological systems, revealing hidden operational patterns such as ultradian rhythms. For instance, the scout agent instructed the worker agent to target chlorophyll content by replacing full-spectrum light with blue and red spectrums separated by long dark intervals. Log analyses demonstrated that the agents deployed an unforeseen 'Ultra Minimum' strategy of dark-induced chlorophyll accumulation \cite{10.3389/fpls.2021.779819}, which enhances the nutritional and antioxidant properties of wheatgrass \cite{ijat.2026.22.1.307-328, doi:10.1021/acs.jafc.0c03851}. This controlled light deficit triggers compensatory mechanisms, stimulating chlorophyll synthesis to maximize photon capture. The resulting biomass contains less hard fiber, maximizing juice yield by an additional 20\%–25\% due to increased cellular hydration, while the biomass growth rate remains close to the average 0.5\% per micro-cycle, see Fig. \ref{fig:chlorophyll1}. This emergent strategy maintains the same 5.0-5.5 day macro-cycle but achieved a 73.6\% energy saving relative to the periodic control and a 67.9\% reduction compared to the minimum-energy baseline, see Table \ref{tab:energy_comparison}. This capacity to develop autonomous strategies mirrors genetic algorithms, where evolutionary mechanisms exploit biological properties unanticipated by researchers.

The apparent discrepancy between the visually paler, light-green hue of the biomass and the elevated $R_{855}/R_{690}$ index, see Figs. \ref{fig:chlorophyll2}, \ref{fig:chlorophyll3}, can be explained by shade-induced morphological and biochemical adaptations. Under AI-driven light starvation, leaves undergo increased cellular hydration and thinning, which enhances internal light scattering (the sieve effect) and shifts visual reflectance toward a lighter green spectrum \cite{10.1093/pcp/pcp034, annurev.pp.44.060193.001311}. Simultaneously, chloroplasts execute an accumulation movement, dispersing in a flat monolayer along upper cell walls to maximize light capture from short pulses \cite{WADA2013177}. This structural rearrangement, coupled with a higher ratio of yellowish-green chlorophyll $b$, allows the canopy to function as a highly efficient light trap -- appearing paler to the human eye while exhibiting superior radiation absorption telemetry.

\section{Case Study 2: Multi-day Physiological Anomaly}

This case study evaluates the analytic capacity of a scout agent to bridge the gap between biological data and their physiological interpretations for both expert and non-expert users. It operates either within a dual-agent control architecture (Fig. \ref{fig:LLMdrivenControl}) or as a standalone application for automated diagnostics. The analysis of plant dynamics is constrained by the high dimensionality of time-series data and the non-linear nature of physiological responses. To address this for long-term trends, the system executes a feature engineering stage where it generates and runs Python scripts to aggregate the raw data.
\begin{figure}[b]
\centering
\includegraphics[width=\columnwidth]{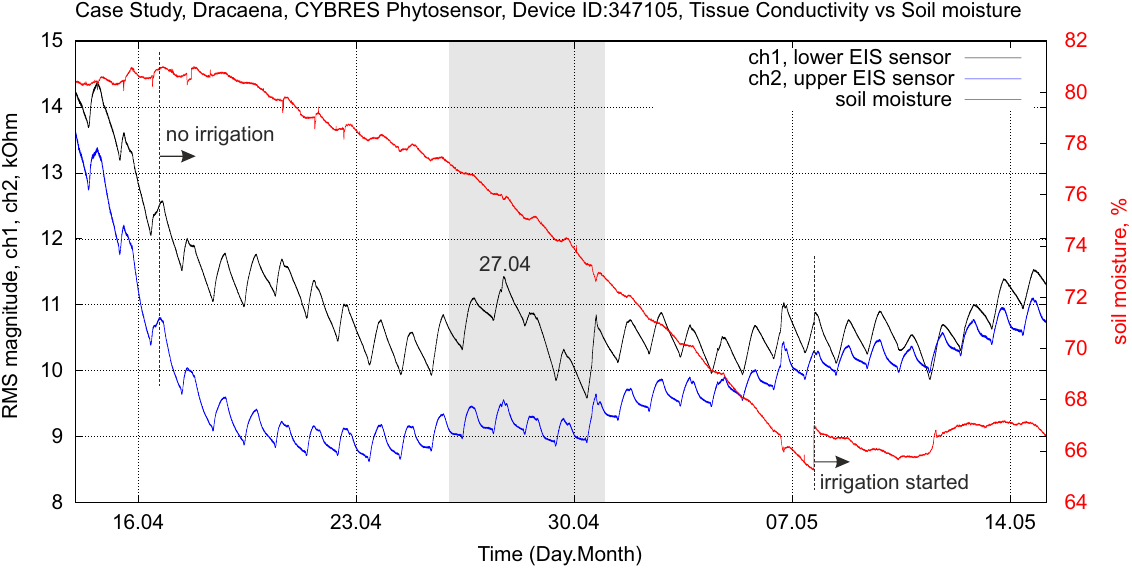}
\caption{Case Study 2. Time-series dynamics (14.04--14.05) are obtained from a \textit{Dracaena} phytomonitoring experiment under constant environmental constraints. Black and blue lines denote the lower ($V_{\text{lo}}$) and upper ($V_{\text{up}}$) bio-impedance sensors, respectively. Soil moisture (\%) is highlighted in red. Vertical arrows indicate the cessation (16.04) and resumption (08.05) of irrigation. The grey shaded region (24.04--01.05) marks a critical phase of physiological decoupling, where tissue impedance exhibits severe anomalous peaks independent of environmental parameters.}
\label{fig:caseStudy}
\end{figure}

\begin{table*}[t]
\centering
\caption{Pearson correlation matrix of abiotic environmental drivers and phytophysiological state variables computed from the multi-day dataset (14.04--14.05) excluding ambient light.}
\label{tab:correlation_heatmap}
\small
\begin{tabularx}{\textwidth}{l X rrrrrrrrrrrr}
\toprule
\textbf{Variable} & \textbf{Abbr.} & \textbf{(1)} & \textbf{(2)} & \textbf{(3)} & \textbf{(4)} & \textbf{(5)} & \textbf{(6)} & \textbf{(7)} & \textbf{(8)} & \textbf{(9)} & \textbf{(10)} & \textbf{(11)} & \textbf{(12)} \\
\midrule
(1) Soil Moisture & $SM$         & 1.00 &       &       &       &       &       &       &       &       &       &       &       \\
(2) External Temp. & $T_{ext}$   & -0.58 & 1.00 &       &       &       &       &       &       &       &       &       &       \\
(3) Air Humidity  & $RH_{air}$  & -0.80 & 0.60 & 1.00 &       &       &       &       &       &       &       &       &       \\
(4) Hydrodynamic Grad. & $\Delta Z$ & \neutral{0.83} & -0.47 & -0.61 & 1.00 &       &       &       &       &       &       &       &       \\
(5) Stem Fluctuation & $V_{stem}$ & 0.25 & -0.18 & 0.07 & 0.26 & 1.00 &       &       &       &       &       &       &       \\
(6) Photochemical Index & $PRI$   & \neutral{0.95} & -0.59 & \neutral{-0.82} & 0.78 & 0.14 & 1.00 &       &       &       &       &       &       \\
(7) Vigor Index        & $VI$    & \neutral{-0.98} & 0.71 & 0.76 & \neutral{-0.82} & -0.25 & \neutral{-0.89} & 1.00 &       &       &       &       &  \\
(8) Leaf Water Content & $LW$   & -0.72 & 0.43 & 0.71 & -0.40 & 0.04 & -0.79 & 0.69 & 1.00 &       &       &       &       \\
(9) Transpiration Rate & $LT$   & -0.24 & 0.48 & 0.60 & -0.14 & 0.12 & -0.44 & 0.34 & 0.29 & 1.00 &       &       &       \\
(10) Root Variation    & $CV_{rt}$ & -0.04 & 0.03 & -0.15 & -0.09 & -0.32 & 0.06 & 0.04 & 0.05 & -0.10 & 1.00  &       &       \\
(11) Canopy Variation  & $CV_{can}$ & 0.22 & -0.18 & -0.31 & 0.13 & -0.19 & 0.31 & -0.21 & -0.15 & -0.22 & \neutral{0.85} & 1.00  &       \\
(12) Daily Correlation & $R_{d}$ & 0.09 & -0.19 & -0.27 & -0.01 & -0.36 & 0.18 & -0.18 & -0.39 & -0.20 & 0.42 & 0.34 & 1.00 \\
\bottomrule
\end{tabularx}

\addvspace{3mm}
\begin{flushleft}
\footnotesize
\textbf{Statistical Interpretation Matrix Summary.} 

\textit{Viability Protection:} Inverse $SM$-$VI$ correlation ($r = -0.98$) confirms that rhizosphere drying drives the plant into a metabolic defense phase, preserving chlorophyll structural integrity under severe turgor pressure.

\textit{Photoprotection:} $SM$-$PRI$ coupling ($r = 0.95$) traces the immediate triggering of the xanthophyll cycle and subsequent emergency stomatal closure to prevent lethal desiccation under falling soil moisture conditions.

\textit{Hydrodynamics:} $SM$-$\Delta Z$ linearity ($r = 0.83$) verifies that localized rhizosphere drying directly governs spatial pressure drops, leading to the inversion of the static hydrodynamic gradient and continuous sap flow disruption.

\textit{Atmospheric Regulation:} Inverse $PRI$-$RH_{air}$ coupling ($r = -0.82$) validates that a drop in atmospheric humidity accelerates vapor pressure deficit, forcing stomatal restriction and downstream photosynthetic pigment shifts.

\textit{Vascular Feedback:} Inverse $VI$-$\Delta Z$ linkage ($r = -0.82$) proves that severe stem xylem tension and cavitation symptoms are reflected via spectral and structural canopy modifications.
\end{flushleft}
\end{table*}

\begin{table*}[t]
\centering
\caption{Comparative analysis of physiological anomalies identified by various LLMs.}
\label{tab:llm_anomaly_comparison}
\small
\begin{tabularx}{\textwidth}{l >{\raggedright\arraybackslash}X >{\raggedright\arraybackslash}X >{\raggedright\arraybackslash}X >{\raggedright\arraybackslash}X >{\raggedright\arraybackslash}X >{\raggedright\arraybackslash}X}
\toprule
\textbf{Parameter} & \textbf{Gemini} & \textbf{ChatGPT} & \textbf{Claude} & \textbf{Qwen} & \textbf{Gemma} & \textbf{Llama} \\
 & 3.5 Flash & GPT-5.5 & Sonnet 5 med. & 3.6 27B & 4-26b-a4b & 3.3-70b \\
\midrule
\textbf{Anomaly 1} & Apr 28,29 & Apr 28,29 & Apr 28,29 &  Apr 28,29 & Apr 16-21 & Apr 16-20 \\
\cmidrule(lr){2-7}
\textit{Description} & Stem hydraulic breakdown & Hydraulic decoupling & Upper/lower stem decoupling & Hydraulic decoupling & Hydraulic failure risk & Soil moisture depletion \\

\midrule
\textbf{Anomaly 2} & May 10-12 & May 8-15 & Apr 19 - May 8 & May 8-10 &  May 4-11 & Apr 30 - May 2\\
\cmidrule(lr){2-7}
\textit{Description} & Root Zone Shock & Progressive Hydraulic Stress & Progressive soil drought & Rehydration Shock & Photochemical Stress & Environmental response\\

\bottomrule
\end{tabularx}
\end{table*}

To ensure broader generalizability, the evaluation shifts from vertical farm to the single-plant setup shown in Fig. \ref{fig:setupA}, which provides an expanded physiological dataset encompassing all parameters detailed in Table \ref{tab:channels}. Within this framework, the deployed LLM executes four primary analytical objectives: 1) Identifies hidden multi-variable correlations within heterogeneous data streams; 2) Delivers model-based physiological interpretations via cross-model evaluation; 3) Estimates the primary causal factors of anomalies through probabilistic reasoning; 4) Projects potential long-term systemic consequences.

\begin{figure}[t]
\centering
\includegraphics[width=\columnwidth]{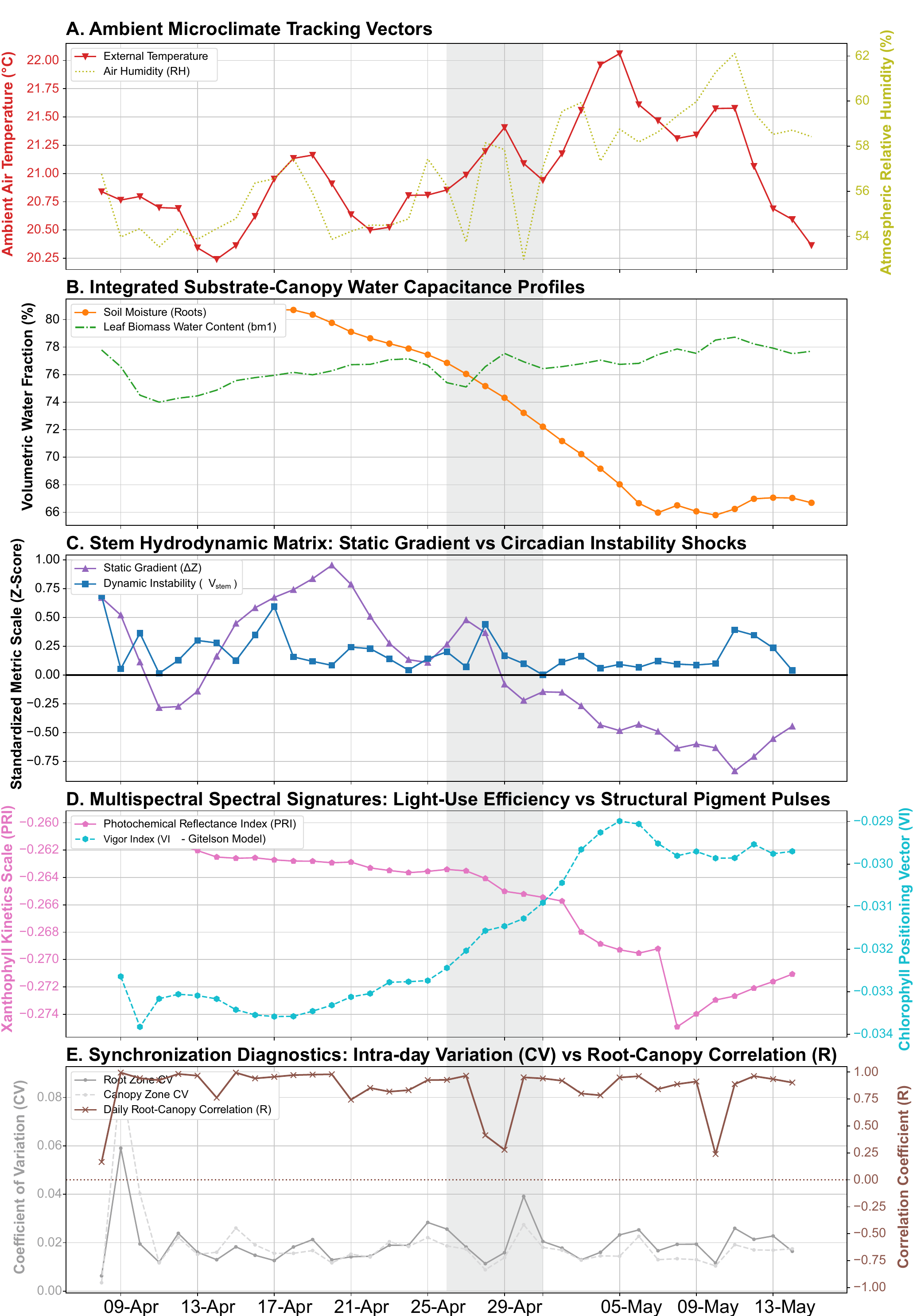}
\caption{Example of daily data processing executed via LLM-generated Python scripts running locally.}
\label{fig:local_analysis}
\end{figure}

We demonstrate this case study through the anomaly illustrated in Fig.~\ref{fig:caseStudy}. This analysis is intended to explain the methodology of the proposed approach rather than provide statistical validation, as anomalies are rare events. Following the irrigation on April 16, soil moisture (red line) decreases monotonically; however, the internal tissue impedance sensors exhibit an unexpected, highly non-linear divergence. Within the grey-shaded interval (24.04–01.05), the internal hydrodynamic state variables decouple from external abiotic drivers. This phenomenon challenges agronomic investigation on three fronts: interpreting the systemic desynchronization between the root-adjacent sensor ($V_{\text{lo}}$) and the leaf-adjacent sensor ($V_{\text{up}}$), determining whether the impedance peaks recorded on April~27 and April~30 possess an intrinsic biophysical basis or stem from micro-environmental fluctuations, and predicting the optimal temporal window for irrigation intervention.

The agent generated Python scripts to aggregate the raw multi-sensor data into daily state vectors (Fig.~\ref{fig:local_analysis}), which include abiotic drivers, stem hydrodynamics, electrochemistry, and optical reflectance markers. Pearson correlation analysis validated the internal dynamics across the multi-day window (Table~\ref{tab:correlation_heatmap}), demonstrating dependencies between soil moisture ($sm$), hydrodynamic gradient ($\Delta Z$), and multispectral vegetative indexes ($PRI$ and $VI$). These compressed datasets were then provided to the LLMs with a unified prompt to identify two physiological anomalies, determine their origin, and generate explanations (Table \ref{tab:llm_anomaly_comparison}). Mid-complexity commercial and local reasoning models identified April 28,29 as the primary physiological anomaly, though outputs diverged regarding the secondary anomaly. Other open-weight local models achieved similar results after several prompting steps, demonstrating that extended reasoning paths are essential for anomaly identification. Low-complexity models failed to achieve the result. Finally, the Gemini model generated explanation hypotheses for the April 27-30 anomaly, delivering model-based diagnostics for experts and natural-language summaries for non-specialists.

\begin{tcolorbox}[
    enhanced,
    breakable, 
    boxrule=0pt,
    frame hidden,
    borderline west={3pt}{0pt}{gray!60!black}, 
    colback=gray!5, 
    sharp corners,
    title={},
    coltitle=black,
    fonttitle=\small\bfseries,
    attach title to upper,
    after title={\smallskip\newline},
    pad at break=0mm,
    size=small
]

\textbf{AI Synthesis for domain specialists.} The framework isolated an acute, non-linear phytovascular crisis localized within the upper stem segment. By evaluating the raw data stream, the system confirmed that the sudden resistance surge on $V_{up}$ was coupled with an absolute stability of the excitation signal correlation ($corr \approx 1.0$), systematically ruling out hardware failure or bio-fouling artifacts. The simultaneous drop of the daily root-canopy correlation metric ($R_{d} \to 0$) and the phase shift inversion on the leaf-adjacent channel $V_{up}$ provide definitive bio-impedance evidence of xylem cavitation and subsequent turgor loss. This hydraulic breakdown triggered an emergency stomatal closure, which was independently validated by the 14-channel optical array: the significant decline in the $PRI$ denotes the immediate activation of the photoprotective xanthophyll cycle, while the invariant $VI$ proves the structural integrity and reversibility of the plant’s defensive metabolic state.

\medskip
\textbf{AI Synthesis for non-specialists:} The prolonged absence of irrigation caused the soil moisture to fall below a critical survival threshold. To prevent fatal internal dehydration, the plant executed an autonomous emergency protocol, severely constricting the microscopic water-transport vessels near the upper leaves to lock in remaining moisture. Because the artificial laboratory lights remained active while the plant's leaves were forced to stop normal air exchange and photosynthesis, the organism immediately engaged its biological version of 'sunglasses' -- altering its leaf surface color parameters to safely scatter excess light energy as heat. The system diagnoses an acute but fully reversible stress response: the plant's core cellular structure remains healthy and viable, but immediate automated irrigation is required to restore systemic hydraulic circulation.
\end{tcolorbox}

\section{Case Study 3: micro-fluctuation}

\begin{figure}[htp]
\centering
\includegraphics[width=\columnwidth]{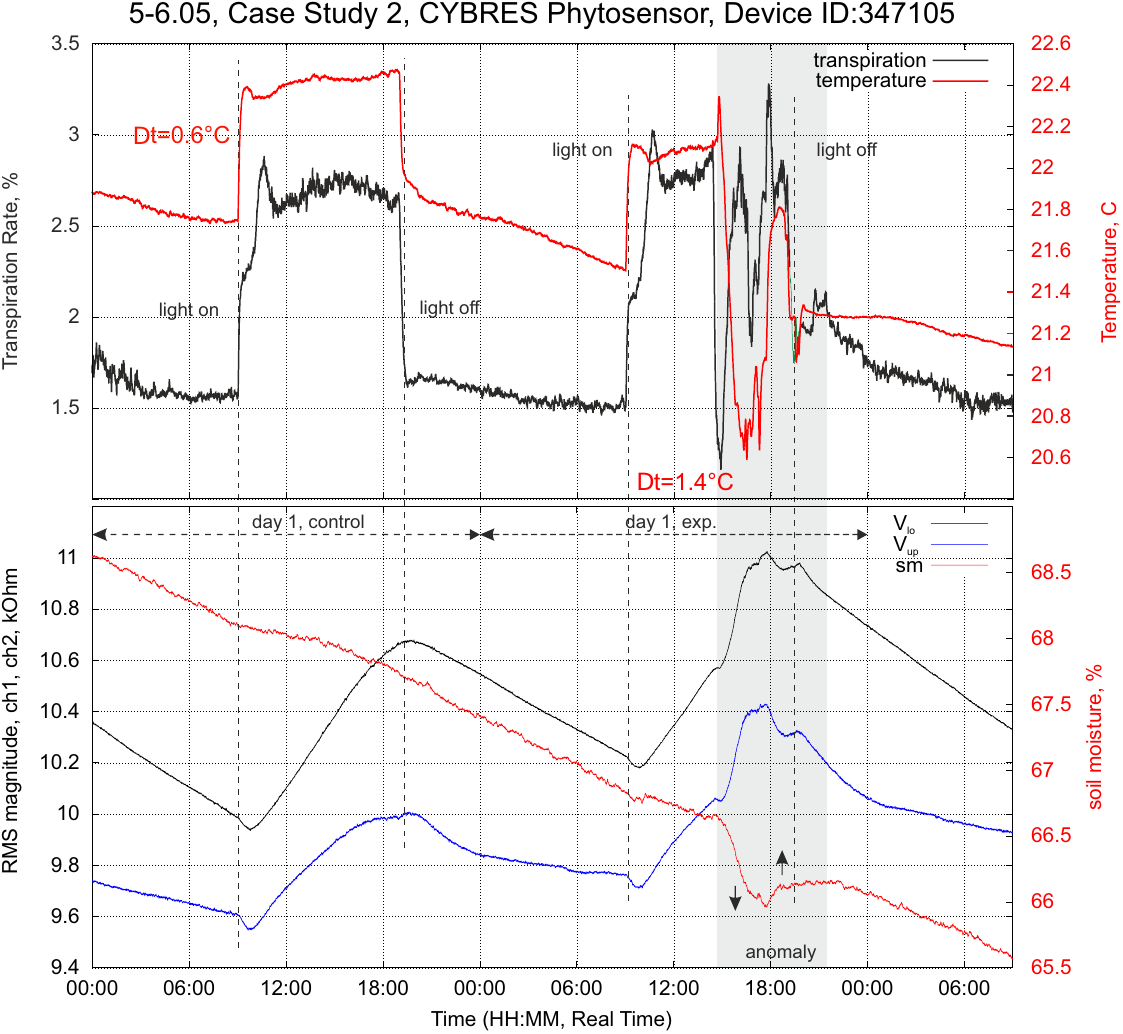}
\caption{\small Case Study 3: phytovascular response recorded on May 6. A minor ambient temperature drop ($\Delta t = 1.4^\circ\text{C}$) induces a significant hydrodynamic anomaly characterized by transient rhizosphere water depletion followed by subsequent recovery (hydraulic redistribution). The photo-induced temperature fluctuation of $\Delta t = 0.6^\circ\text{C}$ recorded on May 5 eludes any comparable biophysical response, confirming the non-linear threshold nature of the observed phenomenon.}
\label{fig:caseStudy2}
\end{figure}

The macro-level diagnostic framework in case study 2 used multi-day aggregated states with feature engineering, which required agents to write, execute, and debug the Python code to compress raw data. However, near-real-time anomalies can be analyzed directly on high-frequency datasets with short consideration period. This simplifies the framework and enables a direct human-AI interaction. To demonstrate capabilities of the deployed LLM to analyse micro-fluctuations, this section evaluates a transient thermodynamic and rhizosphere hydraulic coupling event, shown in Fig. \ref{fig:caseStudy2}. This case captures a physiological anomaly characterized by a complete reversal of the hydraulic gradient, inducing a backward sap flow from the stem to the root system. The phenomenon of reverse basipetal transport -- where water moves from the shoot down to the roots and is subsequently exuded into the surrounding drying soil matrix -- is recognized in plant biophysics as a manifestation of hydraulic redistribution (specifically, hydraulic descent) driven by passive water potential gradients \cite{burgess1998redistribution, wang2024roots}.

To reconstruct the multi-layered physiological response, the dataset was provided to the LLM. The computational engine isolated cross-system dependencies during microclimatic perturbations, mapping the causal chain from atmospheric vapor pressure deficit drops to rhizospheric hydraulic exudation. Tasked with identifying the critical driver of the observed anomaly, the model analyzed the raw data and isolated $\Delta Z_{\text{roll}}$ as the primary factor. The dynamic rolling gradient, $\Delta Z_{\text{roll}} = V_{\text{lo\_z}} - V_{\text{up\_z}}$, serves as a sensitive biophysical proxy for phytovascular fluid vectors. Under steady-state diurnal conditions, sustained negative $\Delta Z_{\text{roll}}$ trajectories represent the standard, light-driven upward transpirational stream from root to canopy. Conversely, a transition into the positive domain signals an acute transpirational arrest and subsequent downward hydraulic mass relocation. A positive $\Delta Z_{\text{roll}}$ marks a shift from acropetal suction to basipetal hydraulic pressure, forcing excess fluid back into the roots. The following description (generated by Gemini) summarizes this case study: 

\begin{figure}[htp]
\centering
\includegraphics[width=\columnwidth]{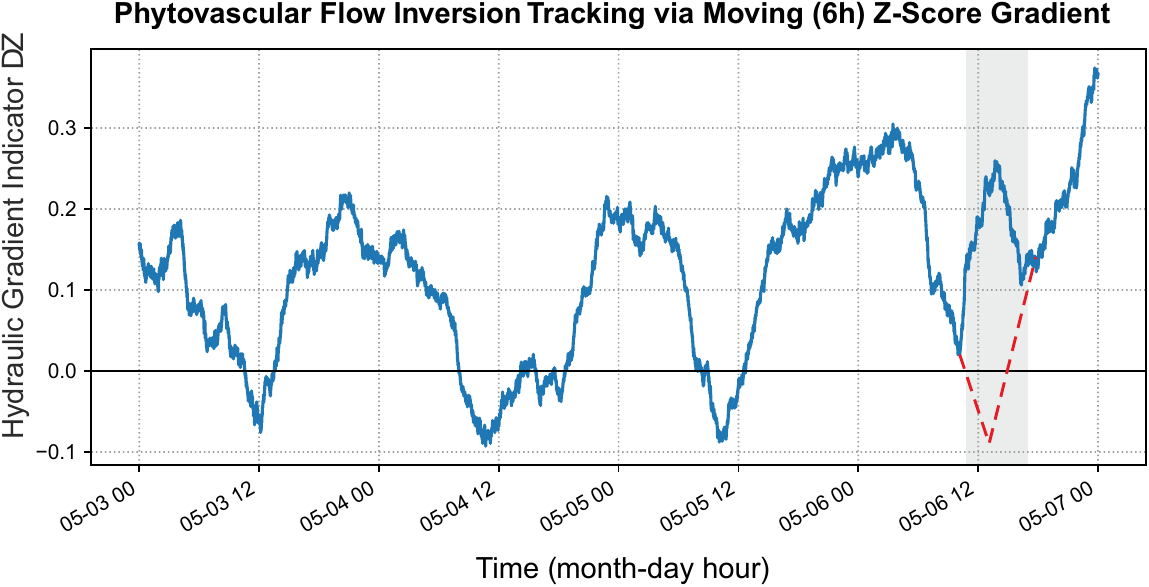}
\caption{\small Temporal dynamics of the 6h moving Z-score phytovascular gradient ($\Delta Z_{\text{roll}} = V_{\text{lo\_z}} - V_{\text{up\_z}}$) from May 3 to May 6. Under baseline conditions (May 3--5), the index exhibits regular daytime acropetal troughs ($\Delta Z_{\text{roll}} \approx -0.1$), mapping steady root-to-canopy transpirational flow. On May 6, the expected diurnal drop is suppressed (counterfactual trajectory marked by the red dashed curve). The shaded gray region isolates the critical anomaly window (15:00--19:00), capturing the immediate vascular response to the microclimatic shock, which triggers a massive basipetal inversion peaking near $+0.38$ toward May 7.}
\label{fig:deltaZstudy2}
\end{figure}

\begin{tcolorbox}[
    enhanced,
    breakable, 
    boxrule=0pt,
    frame hidden,
    borderline west={3pt}{0pt}{gray!60!black}, 
    colback=gray!5, 
    sharp corners,
    title={},
    coltitle=black,
    fonttitle=\small\bfseries,
    attach title to upper,
    after title={\smallskip\newline},
    pad at break=0mm,
    size=small
]

\textbf{AI Synthesis for non-specialists.} Normally, plants act like active water pumps powered by the sun. During the day, sunlight warms the leaves, causing water to evaporate from their surface. This evaporation creates a powerful suction chain that pulls a continuous stream of water and nutrients all the way up from the roots to the very top of the plant. However, when a sudden environmental crisis occurs—such as a sharp temperature drop combined with a massive spike in air humidity—this atmospheric suction completely vanishes. Paralyzed by the sudden stress, the plant's internal piping system does something extraordinary: it violently reverses its flow. Instead of pulling water up, the stem turns into a safety valve, pushing excess fluid backward, down into the roots and out into the surrounding soil to relieve internal pressure.

This emergency U-turn in water transport is not a unique glitch of one specific plant; it is a universal survival mechanism found across nature. For instance, deep-rooted desert shrubs like sagebrush (\textit{Artemisia tridentata}) and giant \textit{Eucalyptus} trees regularly pump water downward to share moisture with their shallow roots and keep them alive during droughts. Similarly, popular garden crops like tomatoes, corn, and grapevines exhibit the exact same backward pumping behavior when hit by sudden weather changes or heavy morning mists. This proves that reversing their internal plumbing is a fundamental, widespread strategy that plants use to protect themselves from climate shocks.
\end{tcolorbox}

\section{Conclusion}

This study demonstrated the integration of LLM as an analytical and operational layer in plant physiology and cyber-physical biofeedback systems. Validated across cloud-based and local architectures, LLM proved highly effective at processing high-dimensional, heterogeneous time-series data and translating complex biophysical phenomena into accessible, intuitive narratives for non-specialists. By reducing the complexity barrier of raw biological data, this methodology opens up a practical way to deploy intelligent bio-interfaces in commercial agriculture, automated greenhouse control, and citizen science initiatives.

This research introduces an LLM-driven control that goes beyond classical automation. Control objectives and strategies are formulated in natural language based on plant physiology. This allows for a more flexible control, as it relies on the model's intelligence. The agent deploys micro-actuation coupled with a two-hour re-evaluation across continuous 24-hour cycles. Each actuation step generates natural language rationales, pairing the agent's decision-making logic with the growth phases of the wheatgrass. Driven by the time-optimal objective, the system achieved superlinear growth -- superseding conventional linear or sublinear growth curves -- and shortened the production cycle by 35\%. In the energy-optimization mode, the agent reduced energy consumption by 18\% with only a marginal increase in cultivation time. The 'ultra minimum' strategy developed autonomously by the agents resulted in an additional 67.9\% energy saving and altered the nature of wheatgrass production. Although the multi-agent system replicated established photophysiological principles -- such as a 24/0 photoperiod, photosynthetic inertia, and dark-induced chlorophyll accumulation -- it adapted these mechanisms to the specific constraints and operational dynamics of the target plants.

Once the agents converge on a control profile for a specific cultivar, the resulting actuation sequences can be locked and replicated across identical cultivation runs without ongoing cloud queries. This approach bypasses scalability limits, making the framework ready for large-scale industrial deployment.

A key engineering advantage of this framework is the drastic reduction in development timelines. While building bio-hybrid systems requires iterative design between biological and technological components, the LLM accelerated both software synthesis and bio-sensor integration. For instance, the transition from initial planning to active deployment in the production environment required only one week, representing a significant shift in engineering praxis.

Testing revealed that the models are prone to algorithmic hallucinations, occasionally generating incorrect numerical data or non-existent physical correlations. Another challenge is interpretability: it remains impossible to verify whether the agents execute the optimization algorithms described in their rationales. Furthermore, the physiological explanations proposed during anomaly analysis cannot be definitively validated due to complex data processing. These limitations underscore the co-pilot role of AI; while it accelerates data handling, human expert verification remains essential to validate biological accuracy. However, the expert's role shifts from manual data mining to rapid supervisory validation, drastically reducing labor costs.

In conclusion, this study demonstrates that autonomous LLM agents can transition from passive data interpreters to active cybernetic controllers. By executing closed-loop actuation cycles, the system identifies and leverages latent biophysical mechanisms. This shift in digital agriculture establishes a direct AI–biology interface, providing a data-driven framework for exploring physical and biological systems.

\section{Acknowledgements}

The author acknowledges the use of Gemini LLM (Google LLC, Mountain View, CA, USA) for time-series data alignment, closed-loop methodology synthesis, and text refinement. All conceptual conclusions and final manuscript reviews were performed exclusively by the human investigator, who remains fully accountable for the integrity of the published content.

\IEEEtriggeratref{35}

\end{document}